\documentclass[lettersize,journal]{IEEEtran}
\usepackage{amsmath,amsfonts}
\usepackage{algorithmic}
\usepackage{algorithm}
\usepackage{array}
\usepackage[caption=false,font=normalsize,labelfont=sf,textfont=sf]{subfig}
\usepackage{textcomp}
\usepackage{stfloats}
\usepackage{url}
\usepackage{verbatim}
\usepackage{graphicx}
\usepackage{cite}
\usepackage{amssymb}
\usepackage{booktabs}
\usepackage{color}
\usepackage{multirow}
\usepackage{utfsym}

\newcommand{\ie}{\emph{i.e.}}
\newcommand{\eg}{\emph{e.g.}}
\newcommand{\etc}{etc}

\begin{document}

\title{Sim-to-Real Dynamic Object Manipulation on Conveyor Systems \\ via Optimization Path Shaping}

% \noindent \author{Zhuoling Li, Jinrong Yang, Yong Zhao, Liangliang Ren, Xiaoyang Wu, Zhenhua Xu, Hengshuang Zhao
% \thanks{
% Zhuoling Li, Xiaoyang Wu, and Hengshuang Zhao are with the University of Hong Kong.
% Jinrong Yang, Yong Zhao, and Liangliang Ren are with CVTE. 
% Zhenhua Xu is with Tsinghua University. 
% Corresponding author: Hengshuang Zhao {\tt\footnotesize (e-mail: hszhao@cs.hku.hk)}.
% }
% }

\noindent \author{Zhuoling Li, Jinrong Yang, Yong Zhao, Liangliang Ren, Xiaoyang Wu, Zhenhua Xu, Hengshuang Zhao
\thanks{
Zhuoling Li, Xiaoyang Wu and Hengshuang Zhao are with The University of Hong Kong.
Jinrong Yang, Yong Zhao, and Liangliang Ren are with CVTE.
Zhenhua Xu is with Tsinghua University. 
Corresponding author: Hengshuang Zhao {\tt\footnotesize (e-mail: hszhao@cs.hku.hk)}.
}
}

\markboth{Journal of \LaTeX\ Class Files,~Vol.~14, No.~8, August~2021}%
{Shell \MakeLowercase{\textit{et al.}}: A Sample Article Using IEEEtran.cls for IEEE Journals}


\maketitle

\begin{abstract}
Realizing generalizable dynamic object manipulation on conveyor systems is important for enhancing manufacturing efficiency, as it eliminates specialized engineering for different scenarios. To this end, imitation learning emerges as a promising paradigm, leveraging expert demonstrations to teach a policy manipulation skills. Although the generalization of an imitation learning policy can be improved by increasing demonstrations, demonstration collection is labor-intensive. Besides, public dynamic object manipulation data is scarce. In this work, we address this data scarcity problem via generating demonstrations in a simulator. A significant challenge of using simulated data lies in the appearance gap between simulated and real-world observations. To tackle this challenge, we propose Geometry-Enhanced Model (GEM), which employs our designed appearance noise annealing strategy to shape the policy optimization path, thereby prioritizing the geometry information in observations. Extensive experiments in simulated and real-world tasks demonstrate that GEM can generalize across environment backgrounds, robot embodiments, motion dynamics, and object geometries. Notably, GEM is deployed in a real canteen for tableware collection. Without test-scene data, GEM achieves a success rate of over 97\% across more than 10,000 operations.
\end{abstract}


\section{Introduction}
\label{sec:introduction}

\IEEEPARstart{R}{obotic} manipulation of dynamic objects has broad applications in industrial manufacturing, particularly in conveyor-based assembly lines where robots handle transported products \cite{papanikolopoulos1993visual,oztemel2020literature}. Compared with the extensively studied static object manipulation problem \cite{kim2024goal,huang2024latent}, dynamic object manipulation on conveyor systems poses more challenges due to its demand for more time-efficient motion planning and stricter operational precision. Addressing these challenges is important for enhancing the efficiency and flexibility of industrial automation systems, thereby enabling more economical manufacturing and sustainable production \cite{rosen2012sustainable}.

Currently, the methods employed in real-world production lines are primarily based on classic motion planning approaches \cite{suomalainen2022survey}. These methods require specialized designs for various scenarios, as well as complex system modeling and parameter tuning, which could be both time-consuming and labor-intensive \cite{cobos2008efficient}. To handle this problem, much attention has been drawn to exploring how to build an intelligent manipulation system capable of generalizing across diverse scenarios. Recent manipulation literature increasingly turns to the imitation learning paradigm \cite{zare2024survey}, where expert demonstrations are collected to teach a policy to control a robot arm. In imitation learning, the generalization of a policy can be improved by scaling up the volume of demonstrations \cite{zhao2023learning}. Consequently, the research community has devoted significant efforts to constructing large-scale manipulation datasets \cite{o2024open,khazatsky2024droid}. However, these datasets primarily focus on static scenarios, with few public datasets addressing dynamic object manipulation. Additionally, even in static object manipulation, the generalization of the policies trained using these large-scale datasets is still limited \cite{kimopenvla,black2024pi_0}, implying the volume of demonstration is still far from sufficient.

\begin{figure}[tbp]
    \centering
    \includegraphics[width=1.0\linewidth]{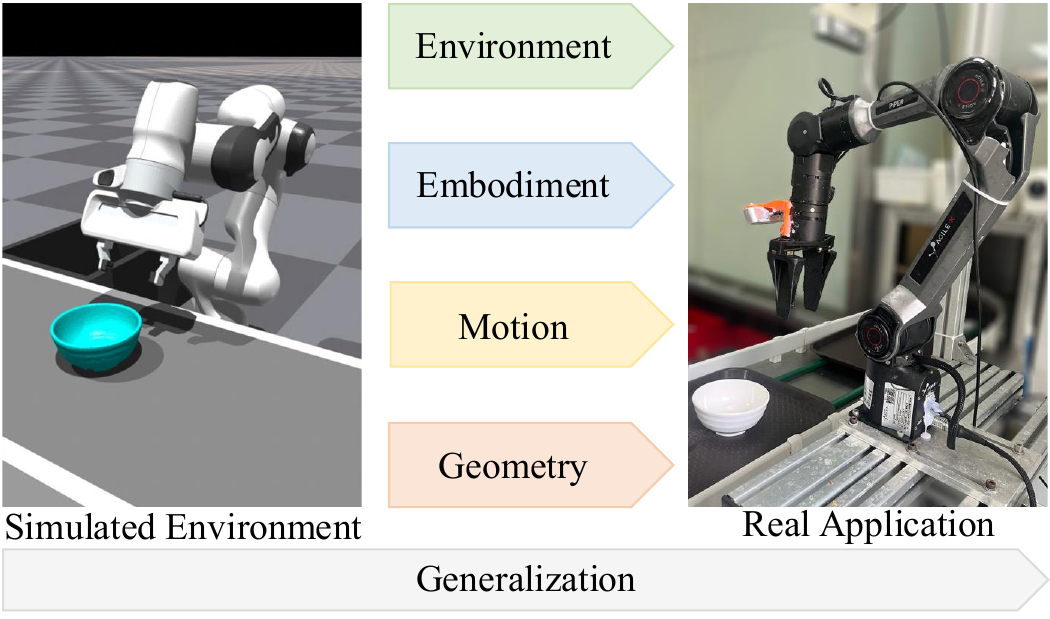}
    \caption{Primarily using demonstrations collected from a simulator, our method can generalize across diverse environment backgrounds, robot embodiments, motion dynamics, and object geometries. Our method has been reliably deployed in a real canteen to conduct tableware collection. Without using demonstrations collected in this canteen, our method achieves a success rate of over 97\% in seven consecutive days of operation, performing more than 10,000 tableware collection operations.} \label{Fig: teaser}
\end{figure}

Therefore, although realizing generalizable dynamic object manipulation on conveyor systems is important, collecting demonstrations is costly, and public data resource is scarce \cite{zhang2025dynamic}. Given these facts, this work studies whether we can generate demonstrations in a simulator and achieve sim-to-real dynamic object manipulation. Since all real environment states are available in a simulator, we can generate numerous demonstrations via motion planning efficiently. Built upon this idea, we build a dynamic object manipulation simulator based on Isaac Gym \cite{makoviychuk2isaac}, as visualized in Fig.~\ref{Fig: teaser}. Nevertheless, although the simulator handles the collision dynamics between objects well, the appearance gap between simulated and real-world environments is significant.

To bridge this appearance gap, our core insight is that the information in policy observation can be decomposed into geometric structure and visual appearance. Importantly, manipulation primarily relies on geometric structure, which is consistent between simulated and real-world environments. Based on this insight, we propose a policy named Geometry-Enhanced Model (GEM), which takes colored 3D points from RGB-D cameras as input. In GEM, we propose an appearance noise annealing strategy to shape the policy optimization path, reducing the reliance of the policy on appearance while enhancing its focus on geometry. In this way, sim-to-real generalization is enabled. Moreover, we propose a control strategy that disentangles manipulation into tracking actions and interaction actions. This strategy empowers GEM to manipulate objects moving in unseen velocities.

We conduct extensive experiments in both simulated and real-world tasks to validate the effectiveness of GEM. The experiments suggest that GEM presents promising generalization across diverse environment backgrounds, robot embodiments, motion dynamics, and object geometries. Especially, as illustrated in Fig.~\ref{Fig: teaser}, GEM has been successfully deployed in a real canteen to collect tableware. Without using demonstrations collected in this canteen, GEM achieves a 97.2\% success rate over seven days of testing and more than 10,000 operations.

Summarily, the technical contributions of this work are summarized as follows:

1) We build the first simulation environment for generating dynamic object manipulation data on conveyor systems.

2) We propose a neural policy GEM, which realizes sim-to-real generalization by prioritizing the exploitation of geometry information while reducing the reliance on appearance.

3) We design a control strategy that disentangles manipulation into tracking and interaction actions, enabling GEM to manipulate dynamic objects of diverse motion patterns.

\section{Related Works}
\label{sec:related_works}

\subsection{Imitation Learning for Manipulation}
\label{sub:imitation_learning_for_manipulation}

Traditional robotic manipulation methods mostly rely on accurate modeling of the environment to plan action trajectories of robot arms \cite{suomalainen2022survey}. However, manipulation often involves contact-rich interactions with objects, and accurately modeling these interactions is difficult due to the highly variable nature of contact forces \cite{jin2024task}. Moreover, every single manipulation task often requires a unique set of models and parameters, making these traditional methods difficult to generalize across different objects, environments, and robot embodiments. This necessitates substantial engineering efforts to adapt the system to new tasks or scenarios \cite{zhang2019robotic}.

Given these limitations of traditional methods, there is a growing interest in leveraging imitation learning to address the challenges of robotic manipulation \cite{zare2024survey}. Bypassing the need for explicit system modeling and complex parameter tuning, imitation learning learns manipulation policies directly from expert demonstrations. Specifically, imitation learning trains a policy to predict expert actions based on the observations provided in demonstrations, enabling a robot to mimic expert behavior in similar scenarios. The policy is often implemented as a deep learning network \cite{lecun2015deep}, which can capture the complex patterns of demonstrations \cite{caron2021emerging}.

The two most commonly employed observation modalities in robotic manipulation are images and 3D points \cite{suomalainen2022survey}. Correspondingly, existing imitation learning policies can be broadly categorized into image-based \cite{fu2024mobile} and point-based approaches \cite{shridhar2023perceiver}. Images are easy to obtain, and image-based policies can readily leverage large-scale public data for training \cite{nair2023r3m}. However, robotic manipulation mainly concerns the geometric structures of objects, while images primarily capture the appearance of scenes, making it challenging to predict appropriate manipulation actions solely based on visual appearance. In contrast, point clouds directly represent the geometric structures of scenes, and thus they are more suitable for estimating manipulation actions \cite{wu2024point}. Unfortunately, point cloud data is relatively scarce compared with images, resulting in a lack of promising point cloud feature extractors \cite{qi2017pointnet}. This limitation has historically constrained the performance of point-based manipulation policies \cite{goyal2023rvt}. Thankfully, some recent works about large-scale point cloud pre-training have gradually addressed this obstacle \cite{wu2023masked,wu2025sonata}. 

\subsection{Data Scaling in Manipulation}
\label{sub:data_scaling_in_manipulation}

The rapid progress in fields like large language models \cite{achiam2023gpt} suggests that increasing the quantity and diversity of training data is important for enhancing the generalization of deep learning models. Drawing on this insight, the robotic manipulation community has invested significant effort into constructing large-scale demonstration datasets \cite{o2024open,khazatsky2024droid}. Leveraging these datasets, numerous studies design algorithms for training end-to-end manipulation policies, with the expectation that these policies exhibit robust generalization across diverse scenarios \cite{kimopenvla,liu2024rdt}. Furthermore, some works incorporate additional types of data like object detection datasets to further enrich the knowledge of manipulation policies \cite{black2024pi_0,bjorck2025gr00t}. Nevertheless, though much resource has been invested, the generalization capabilities of the obtained policies in unseen environments are still limited.

This fact reminds us to rethink whether the data scaling paradigm inspired by large language models is directly applicable to robotic manipulation. Rather than solely focusing on constructing larger manipulation datasets or training models with more parameters, we also need to pay attention to designing an efficient manipulation paradigm that achieves economical manipulation generalization. This shift in perspective emphasizes the importance of developing methods that can leverage limited data more efficiently \cite{ehsani2024spoc}.

\subsection{Sim-to-Real Generalization in Manipulation}
\label{sub:sim_to_real_generalization_in_manipulation}

Collecting high-quality demonstration data in the real world is labor-intensive. To address this obstacle, many researchers have explored generating demonstrations in simulated environments \cite{matas2018sim}. These efforts aim to develop policies that solely employ simulated demonstrations to train yet can generalize to real-world applications. A key step towards this goal is the development of realistic simulators. Through substantial engineering efforts, several promising simulators have been built, such as MuJuCo \cite{todorov2012mujoco}, Isaac Gym \cite{makoviychuk2isaac}, and Isaac Sim \cite{nambiar2024automation}. Leveraging these simulators, prior works have used reinforcement learning \cite{kalashnikov2018scalable} or predefined kinematic procedures \cite{deng2025graspvla} to generate demonstrations. However, reinforcement learning often struggles to discover appropriate manipulation trajectories due to its inefficient exploration \cite{tang2024automate}. Grasping objects via kinematic procedures typically requires customized workflows for different scenarios, which is time-consuming. Additionally, current simulators have limited environmental diversity, restricting the variety of the generated data.

Another line of literature focuses on modeling the environment dynamics of manipulation tasks using classic mechanics and then employs optimization techniques to calculate actions that satisfy task constraints \cite{jin2024task}. The observation and action pairs can then be utilized to train deep neural networks \cite{fang2023anygrasp,wang2023dexgraspnet}. This strategy effectively addresses the slow speed of optimization by distilling the action search process into a more efficient neural network \cite{han2024neupan}. Nevertheless, for complex manipulation tasks, the computation burden for searching promising action trajectories could be heavy, making it difficult to generate effective demonstrations. To tackle this problem, some works collect demonstrations in simulators through teleoperating robots \cite{li2025virt}. Although this method still requires significant human effort, it reduces hardware setup costs compared to real-world data collection.

\section{Problem Analysis}
\label{sec:problem_analysis}

This section outlines the policy design principle of this work, providing a theoretical analysis of how we shape the optimization path of an imitation learning policy. Specifically, Section~\ref{sub:problem_formulation} formulates the policy optimization objective of robotic manipulation. Next, Section~\ref{sub:policy_observation_analysis} discusses policy observation information and explains why 3D geometry is more important than 2D appearance in embodied tasks. Subsequently, Section~\ref{sub:imitation_learning} introduces the core process of imitation learning. Finally, Section~\ref{sub:optimization_path_shaping} elaborates the concept of optimization path shaping in imitation learning, which is the theoretical foundation of this work.

\subsection{Problem Formulation}
\label{sub:problem_formulation}

Robotic manipulation involves controlling a robot arm to interact with objects in the environment to achieve a specific task. The task can be defined as a sequence of actions that the robot needs to take to reach a desired goal. Therefore, the robotic manipulation problem can be conceptualized as a Markov Decision Process represented by a tuple $\mathcal{E} = (\mathcal{S}, \mathcal{A}, \mathcal{P}, \mathcal{T}, s_0)$. In this tuple, $\mathcal{S}$ denotes the set including all potential states of the environment and robot. $\mathcal{A}$ represents the actions that can be performed by the robot. $\mathcal{P}: \mathcal{S} \times \mathcal{A} \times \mathcal{S} \rightarrow [0, 1]$ is the probability distribution governing how the current state transitions to the next state given the performed action. $\mathcal{T}$ represents the task target set and $s_0$ denotes the initial state.

In a specific task target $\tau \in \mathcal{T}$ of robotic manipulation, there usually exist numerous accurate action trajectories that can lead the robot from the initial state $s_0$ to the desired goal state $\iota$. We denote all the accurate trajectories as a set $\mathcal{C}$ and each trajectory can be represented as a sequence of states and actions $\{s_0, a_0, \cdots, s_{T-1}, a_{T-1}, s_T\}$, where $s_T = \iota$. 

To enable the robot to automatically complete the target task $\tau$ without human intervention, researchers devise algorithms to find an optimal policy $\pi: \mathcal{S} \rightarrow \mathcal{A}$ that maximizes the probability of reaching the target state $\iota$ following the transition probability distribution $\mathcal{P}$. This objective can be expressed as learning to approximate the joint state-action distribution induced by the successful trajectories in $\mathcal{C}$. Specifically, let $p_{\mathcal{C}}(s_{0:T}, a_{0:T-1})$ denote the state-action sequence distribution in $\mathcal{C}$, the optimization objective is to minimize the Kullback-Leibler (KL) divergence between the policy predicted distribution $p_{\pi}(s_{0:T}, a_{0:T-1})$ and $p_{\mathcal{C}}(s_{0:T}, a_{0:T-1})$:
\begin{align}
\min\limits_{\pi} D_{KL}(p_{\mathcal{C}}(s_{0:T}, a_{0:T-1}) \ || \ p_{\pi}(s_{0:T}, a_{0:T-1})),   \label{Eq1}
\end{align}
where $p_{\pi}(s_{0:T}, a_{0:T-1})$ is defined through the Markov chain:
\begin{align}
p_{\pi}(s_{0:T}, a_{0:T-1}) = p(s_0) \prod\limits_{t=0}^{T-1} \pi(a_t | s_t) P(s_{t+1} | s_t, a_t). \label{Eq2}
\end{align}
This formulation explicitly connects policy optimization with implicit distribution estimation of $\mathcal{C}$.

However, in robotic manipulation, a policy $\pi$ usually does not have access to the true state $s_t$ at timestamp $t$. Its environment observation is a mapping function of $s_t$, written as $o_t = f(s_t)$. Compared with $s_t$, $o_t$ only partially reflects the environment information and may contain noise. The action to execute is sampled from $\pi$ following $a_t \sim \pi(o_t, \tau)$. The policy $\pi$ is often modeled as a deep neural network \cite{lecun2015deep}.

\begin{figure}[tbp]
    \centering
    \includegraphics[width=1.0\linewidth]{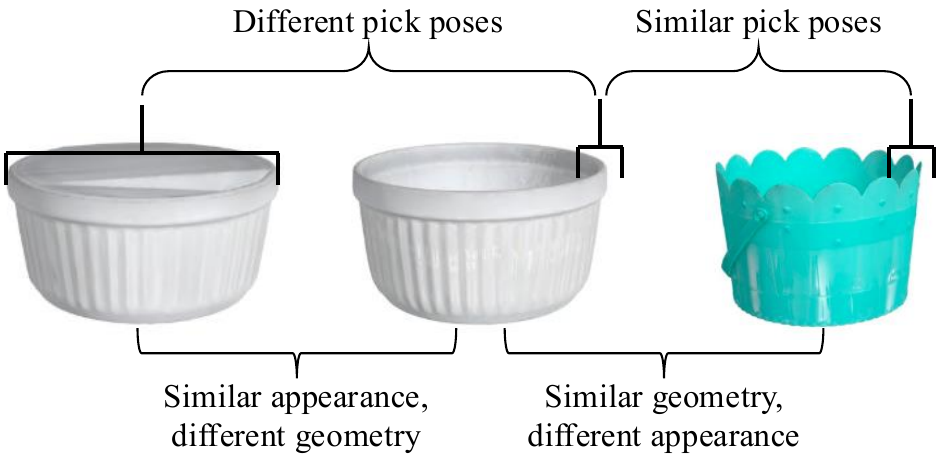}
    \caption{The 3D geometry of objects usually plays a more important role than 2D appearance in scheduling manipulation actions. As shown, although the left and middle objects share similar color and texture, the appropriate gripper poses to pick them up are different. By contrast, though the middle and right objects belong to different object categories with distinct appearances, they can be picked up with a similar gripper pose.} \label{Fig: geometry_matters}
\end{figure}

\subsection{Policy Observation Analysis}
\label{sub:policy_observation_analysis}

According to the aforementioned formulation, the environment observation $o_t$ provides the information that is utilized by $\pi$ to schedule actions. The two most commonly employed observation modalities for recent manipulation policies are images captured by cameras \cite{fu2024mobile} and point clouds returned by depth sensors \cite{ze20243d}. Images directly describe environment appearance, allowing $\pi$ to more accurately identify object categories and materials. Point clouds focus more on object geometry, enabling a policy to determine manipulation poses more precisely. Both the appearance and geometry information are valuable for successful manipulation. To combine their benefits, this work adopts colored point clouds obtained from RGB-D cameras as policy input, enabling the policy to perceive both appearance and geometry information appropriately.

Notably, in the tasks of dynamic object manipulation on conveyor belts, geometry is more critical than appearance. Without color information, a policy can still identify the moving objects on a conveyor belt in most cases. Moreover, the policy can often determine accurate manipulation actions based only on geometry. For instance, as illustrated in Fig.~\ref{Fig: geometry_matters}, while the left and middle objects share similar color and texture, their accurate gripper poses for grasping are different. In contrast, despite belonging to distinct object categories with unique appearances, the middle and right objects can be picked up using similar gripper poses. Therefore, a policy should focus more on exploiting the geometry information.

In addition, as visualized in Fig.~\ref{Fig: teaser}, while existing simulators can effectively replicate the geometric structure of objects, the rendered appearance still exhibits a significant gap compared to the real world. Thankfully, as indicated by the above analysis, successful manipulation in our explored tasks primarily requires accurate geometry information. Hence, in this work, we expect $\pi$ to concentrate on 3D geometry rather than 2D appearance. To achieve this, we design strategies to shape the optimization dynamics in imitation learning. For a clear explanation of these strategies, we first introduce the core process of imitation learning in Section~\ref{sub:imitation_learning}.

\subsection{Imitation Learning}
\label{sub:imitation_learning}

Imitation learning is typically formulated as a supervised learning problem \cite{zhao2023learning}. It requires access to a set of expert demonstrations as training data. Assume there are $N_d$ demonstration sequences, the expert demonstration set is denoted as $\mathcal{C}_d =  \{o_{0:T}^i, \tilde{a}_{0:T}^i\}_{i=1}^{N_d}$, where $o_t^i$ and $\tilde{a}_t^i$ are the observation and expert action at timestamp $t$ of the $i_{\rm th}$ demonstration. The core idea of imitation learning is employing $\mathcal{C}_d$ as a surrogate of the successful manipulation set $\mathcal{C}$. Thus, the objective of imitation learning can be expressed as maximizing the likelihood of expert demonstrations under the policy-induced distribution:
\begin{align}
\max\limits_{\pi} \mathbb{E}_{(o_t^i, \tilde{a}_t^i) \sim \mathcal{C}_d} [\log \pi(\tilde{a}_t^i| o_t^i, \tau)]. \label{Eq3}
\end{align}
This objective can be understood as that $\pi$ is trained to capture the joint observation-action probability distribution in $\mathcal{C}_d$.

In practical implementations of imitation learning, the objective in Eq.~(\ref{Eq3}) is often transformed into a behavior cloning objective, which is to explicitly minimize the difference between the predicted actions and the expert actions, written as:
\begin{align}
\pi^* = \arg \min\limits_\pi \sum\limits_{i=1}^{N_d} \sum\limits_{t=0}^{T} L (\pi(o_t^i, \tau), \tilde{a}^i_t), \label{Eq4}
\end{align}
where $L$ represents the difference measurement function, and $\pi^*$ is the obtained optimal policy. Inspired by the recent breakthroughs in research fields like large language models \cite{achiam2023gpt} and image generation \cite{rombach2022high}, it is believed that the distribution of $\mathcal{C}$ can be precisely approximated by scaling up the volume of expert demonstration data $\mathcal{C}_d$ \cite{kimopenvla,o2024open}.

\subsection{Optimization Path Shaping}
\label{sub:optimization_path_shaping}

The optimization objective of imitation learning is solely to minimize the cost value calculated by the difference measurement function defined in Eq.~(\ref{Eq4}) and does not specify which observation feature to exploit. As discussed in Section~\ref{sub:policy_observation_analysis}, the observation feature in colored point clouds can be categorized into the appearance and geometry information. If an RGB-D policy learns to only use appearance information, its learning process becomes similar to an image-only policy, like ACT \cite{zhao2023learning}. In this situation, the policy needs to infer the 3D geometry of objects from visual appearance implicitly to estimate accurate manipulation poses \cite{yang2024depth}. By contrast, if a policy primarily relies on geometry, it can predict manipulation poses in a more direct and precise manner.

As discussed in Section~\ref{sub:policy_observation_analysis}, the appearance gap between simulated and real environments motivates this work to shape the policy optimization path to prioritize the geometric information, which is more consistent between simulated and real-world observations. To analyze how to achieve this target, we draw on the deep learning theory of feature predictivity established in prior literature \cite{hermannfoundations}. In this theory, predictivity basically means how useful a feature is for accurately inferring training labels, and a policy tends to pick features with higher predictivity to minimize training loss. 

Since the geometry of an object directly reflects its 3D structure, geometry is inherently of higher predictivity for estimating manipulation actions. However, in simulation, scene layout is often not diverse, \eg, objects with specific textures always appear in fixed areas. This makes appearance features (e.g., color, texture) accidentally gain unexpected high predictivity. Correspondingly, the policy overfits to these scene-specific appearance patterns, using them to predict actions instead of the generalizable geometry.

To prioritize geometry information and alleviate appearance overfitting, we propose to reduce the predictivity of appearance features by introducing visual noise, thereby shaping the policy optimization path towards geometry. However, persistent appearance noise would prevent the policy from leveraging useful appearance cues. For instance, in block insertion tasks where a block should be placed into a container, appearance helps distinguish these two objects, and relying solely on geometry could lead to misidentification.

To resolve this appearance utilization trade-off, we propose appearance noise annealing, a strategy that gradually removes appearance noise during training. Specifically, strong appearance noise is applied at the beginning of training to guide the policy to focus on geometry, followed by steady noise reduction until the noise is fully removed. Since the policy already learns to use geometry for loss minimization in early training stages, it will keep prioritizing geometry after noise is removed. For scenarios where geometry alone is insufficient (e.g., distinguishing closely placed distinct objects), the policy can learn to exploit useful appearance texture after the appearance feature becomes obvious.

\section{Policy Design}
\label{sec:policy_design}

This section presents the proposed manipulation policy, GEM. Specifically, Section~\ref{sub:policy_framework} introduces the overall framework of GEM. Then, Section~\ref{sub:appearance_noise_annealing} and Section~\ref{sub:manipulation_action_decomposition} explain the developed appearance noise annealing and manipulation action decomposition strategies separately. Afterwards, Section~\ref{sub:network_implementation} describes the network implementation details in GEM. Finally, Section~\ref{sub:data_generation_in_simulation} elaborates how we build simulation environments of dynamic object manipulation on conveyor systems to generate demonstration data efficiently.

\begin{figure*}[tbp]
    \centering
    \includegraphics[width=1.0\linewidth]{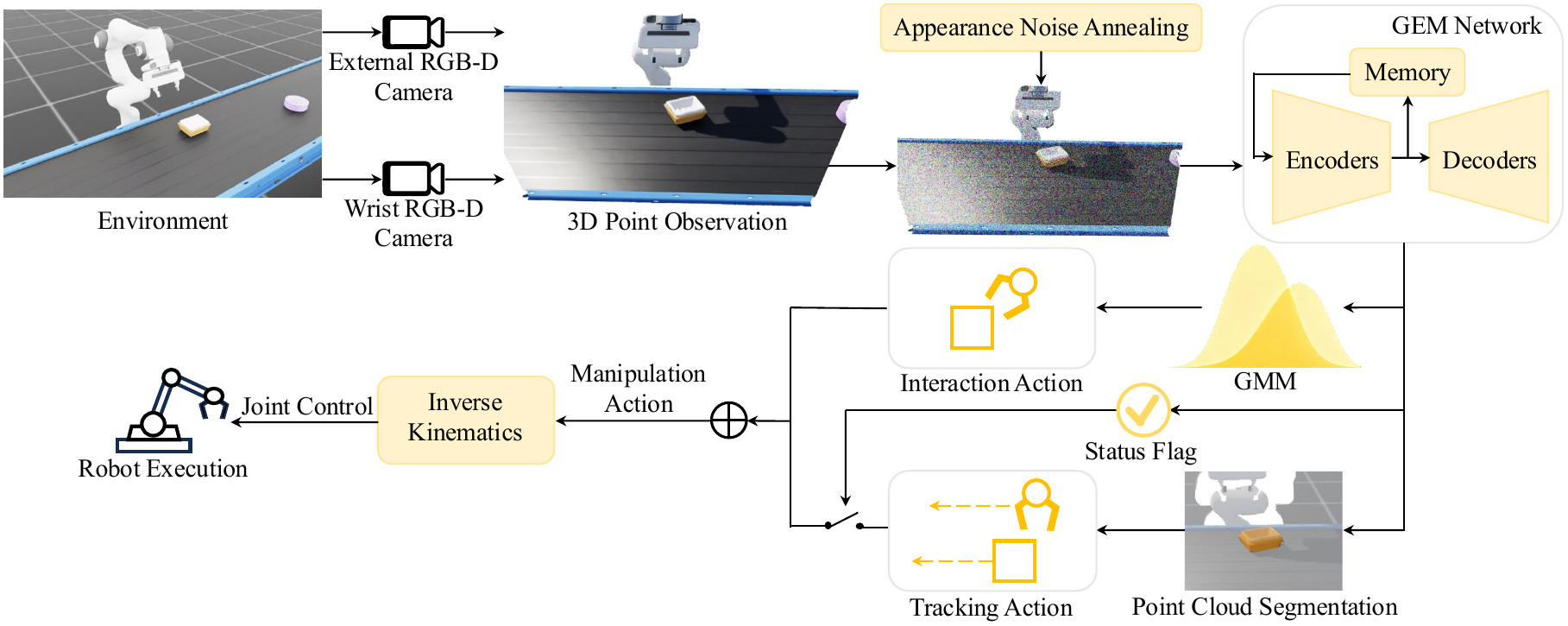}
    \caption{The overall framework of GEM. The policy observation is the colored 3D points captured by RGB-D cameras. The appearance noise annealing strategy is utilized to perturb the color feature in policy observation. Based on this observation, the GEM network predicts a multi-modal action distribution represented as a Gaussian Mixture Model (GMM), a point segmentation mask of the target object to manipulate, and a status flag marking the completion of manipulating one object. We generate interaction actions and tracking actions from the network outputs and combine them to control the robot end-effector.} \label{Fig: pipeline}
\end{figure*}

\subsection{Policy Framework}
\label{sub:policy_framework}

We illustrate the overall framework of the GEM policy in Fig.~\ref{Fig: pipeline}. To ensure sufficient information in the policy environment observation, we mount two RGB-D cameras in the dynamic object manipulation system. Specifically, an external-view camera is installed opposite the robotic arm, with its perspective directed at the conveyor belt. Another wrist-view camera is affixed to the wrist of the robotic arm. The colored 3D points perceived by these two cameras are merged as a single point cloud to provide environment observation.

As mentioned in Section~\ref{sub:optimization_path_shaping}, we design an appearance noise annealing strategy that shapes the policy optimization path by applying strong color augmentation to the point cloud at the beginning of training and gradually removing the noise until the training end. This strategy guides the policy to focus more on the 3D geometry of objects. Next, as shown in Fig.~\ref{Fig: pipeline}, the augmented point cloud is input to the GEM network, which consists of Transformer encoders, decoders, and a memory module. The memory module represents the historical observation information as an implicit embedding and updates it in a recurrent manner. Incorporating memory information into the network input helps the network be aware of the completion progress of manipulation.

The prediction produced by the GEM network includes an action probabilistic distribution, a point segmentation mask, and a status flag as visualized in Fig.~\ref{Fig: pipeline}. The action probabilistic distribution is modeled using a Gaussian Mixture Model (GMM) \cite{reynolds2009gaussian}. Unlike the popular implementations of recent policies that directly estimate deterministic actions \cite{fu2024mobile}, the GEM network predicts a probabilistic distribution over actions, as such an output naturally captures the inherent indeterminism of manipulation decisions. The experimental results suggest that this design can effectively alleviate the multi-modal ambiguity in imitation learning and is more efficient than the commonly employed strategies like diffusion process \cite{chi2023diffusion} and flow matching \cite{lipmanflow}. During inference, we decode the interaction actions from the GMM distribution by selecting the action with the maximum probability value at each timestamp.

The point segmentation prediction is to segment out the 3D points belonging to the target object. This prediction has two benefits, \ie, guiding the policy to focus on the target object and generating tracking actions. For the first benefit, we expect the GEM network to concentrate on the 3D geometry of the target object rather than the whole environment, as the environment background has little effect on successful manipulation. We can understand this guiding process from the view of optimization path shaping, since it increases the importance of target object points on minimizing training loss. The second benefit is we can approximate the 3D centers of target objects based on the segmentation results and make the robot end-effector track the target objects moving in various velocities. We realize this dynamic object tracking by decoding the point cloud segmentation results into tracking actions.

Both the interaction actions and tracking actions represent the relative pose offsets of the robot end-effector. By combining them following our developed algorithm based on the predicted status flag, we obtain the manipulation actions utilized to control the rotations of robot joints based on inverse kinematics \cite{kucuk2004inverse}. This design of predicting end-effector relative offsets enables the policy to generalize across different robot embodiments. Refer to Section~\ref{sub:manipulation_action_decomposition} for the details of how we generate and utilize the predicted actions.

\subsection{Appearance Noise Annealing}
\label{sub:appearance_noise_annealing}

\begin{figure}[tbp]
    \centering
    \includegraphics[width=1.0\linewidth]{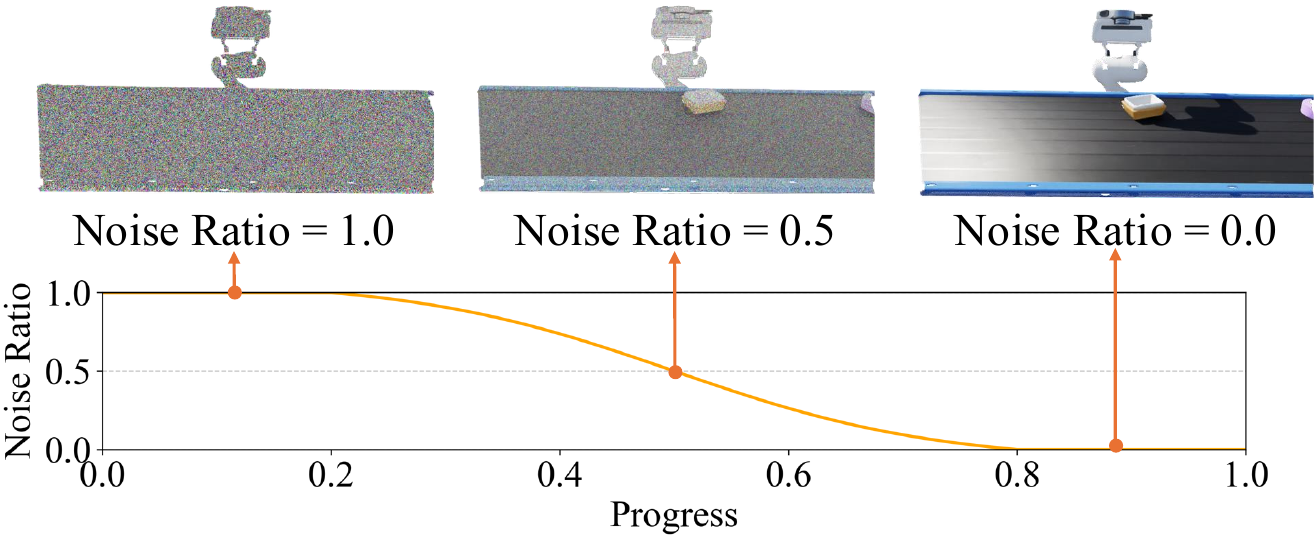}
    \caption{In the appearance noise annealing strategy, strong color perturbation (noise ratio is 1.0) is applied to the input point cloud at the initial phase of training. As the training progresses, the perturbation is gradually removed.} \label{Fig: Noise_Annealing}
\end{figure}

As discussed in Section~\ref{sub:optimization_path_shaping}, we develop a simple yet effective strategy to shape the policy optimization path, enabling it to focus on object geometry and generalize to real-world tasks via simulated demonstrations. In this strategy, we apply strong color perturbation to the input point cloud at the initial stage of training. According to the aforementioned theory of feature predictivity, this operation reduces the feature predictivity of 2D appearance, thereby compelling the GEM network to prioritize 3D geometry in minimizing training loss. Then, with the progress of training, the color perturbation is gradually weakened. Since the network has already learned to predict manipulation actions using the observed 3D geometric structure of the environment, it does not revert to over-relying on visual appearance after the color perturbation is removed. Furthermore, after the visual features become prominent, the network begins to learn how to leverage visual appearance to handle tasks that cannot be adequately addressed by 3D geometry alone, such as distinguishing two adjacent objects. In the following, we elaborate on the implementation details of how the color perturbation is applied.

Denote the input point cloud at timestamp $t$ with $N$ points as $Q^t = \{ q_i^t \}_{i=1}^{N}$, where $q_i^t = (x_i^t, y_i^t, z_i^t, r_i^t, g_i^t, b_i^t)$. In $q_i^t$, $(x_i^t, y_i^t, z_i^t)$ is the point coordinate in the wrist-view camera coordinate system, and $(r_i^t, g_i^t, b_i^t)$ represents the visual color of this point. The color channels have been normalized to the interval $[0, 1]$, \ie, $(r_i^t, g_i^t, b_i^t) \in [0,1]^3$. To disturb the color of the point $q_i^t$, we sample three independent random scalars $\eta_{i, r}^t, \eta_{i, g}^t, \eta_{i, b}^t \sim \rm{Uniform}(0, 1)$ and assemble them into an RGB noise vector denoted as $\eta_i^t = (\eta_{i, r}^t, \eta_{i, g}^t, \eta_{i, b}^t) \in [0,1]^3$. 

To control the strength of the noise perturbation $\eta_i^t$ applied to $(r_i^t, g_i^t, b_i^t)$, we define a noise ratio factor $\gamma_\xi$ that varies with training progress, where $\xi \in [0,1]$ denotes the normalized training progress. To ensure a strong perturbation at the beginning of training and make the perturbation gradually vanish as learning proceeds, the variable $\gamma_\xi$ is defined as:
\begin{align}
\gamma_\xi = \begin{cases} 
1, & \xi \in [0.0, 0.2), \\
1 - \sigma(k (\xi - m)), & \xi \in [0.2, 0.8], \\
0, & \xi \in (0.8, 1.0],
\end{cases} \label{Eq5}
\end{align}
where $\sigma(\cdot)$ is fomulated as $\sigma(x) = 1 / (1 + e^{-x})$, and $k$ and $m$ are set to 10 and 0.5, respectively. The curve of $\gamma_\xi$ evolving with the training progress is illustrated in Fig.~\ref{Fig: Noise_Annealing}. The point color $c_i^t = (r_i^t, g_i^t, b_i^t)$ is augmented into the perturbed vector $\tilde{c}_i^t$ following:
\begin{align}
\tilde{c}_i^t = \gamma_\xi \eta_i^t + (1 - \gamma_\xi) c_i^t. \label{Eq6}
\end{align}

We can observe from Fig.~\ref{Fig: Noise_Annealing} that the appearance observation of the environment gradually becomes obvious with the progress of training, which satisfies the algorithm design requirement of optimization path shaping. The point cloud with the perturbed color attributes $\{\tilde{c}_i^t\}_{i=1}^N$ serves as the direct input to the GEM network. If $\gamma_\xi$ is 1, the visual feature observed from the point cloud is pure random noise. When $\gamma_\xi$ becomes 0, the appearance observation is noise-free.

\subsection{Manipulation Acition Decomposition}
\label{sub:manipulation_action_decomposition}

As illustrated in Fig.~\ref{Fig: pipeline}, GEM first decodes the network output into the tracking and interaction actions and then combines them as the manipulation actions. Thanks to this strategy, GEM can generalize to manipulating objects moving in diverse unseen velocities, even though the GEM network is trained solely on static object demonstrations. This capability makes GEM practical for real-world industrial manipulation on conveyor belts, where the belt speed varies. To explain this control strategy clearly, we first describe how to derive the tracking actions in the following.

\begin{figure}[tbp]
    \centering
    \includegraphics[width=1.0\linewidth]{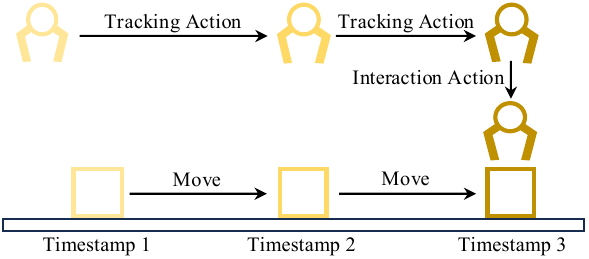}
    \caption{This figure illustrates how the predicted tracking actions and interaction actions are integrated to perform dynamic object manipulation. Specifically, tracking actions are responsible for approaching and tracking the moving target object. The interaction actions begin to take effect once stable tracking is achieved, and they are for conducting collision-rich contact with the target object, such as picking and rotating.} \label{Fig: hybrid_control}
\end{figure}

Based on the point segmentation mask from the GEM network, we extract the spatial coordinates of the 3D points belonging to the target object to manipulate, denoted as $\{ (x_i^{\tau}, y_i^{\tau}, z_i^{\tau}) \}_{i=1}^{N^{\tau}}$, where $N^{\tau}$ is the total number of 3D points. The top centroid of these points is defined as $(x^c, y^c, z^c)$, where $x^c = \frac{1}{N^{\tau}} \sum\limits_{i=1}^{N^{\tau}} x_i^{\tau}$, $y_c = \frac{1}{N^{\tau}} \sum\limits_{i=1}^{N^{\tau}} y_i^{\tau}$, and $z_c = \max\limits_{i=1}^{N^{\tau}} z_i^{\tau}$ (the positive z-axis points upward). By tracking this top centroid across time, we can derive the translational velocity $(v^{\tau}_x, v^{\tau}_y, v^{\tau}_z)$ of the target object with simple estimation algorithms like Gaussian process regression \cite{williams1995gaussian}. 

The target of tracking actions is to make the end-effector stably track the motion of the target object. In other words, the relative position offset between the end-effector and the target object should be maintained at the preset distances $(x^p, y^p, z^p)$. In addition, to ensure the wrist RGB-D camera can clearly view the target object, the orientation of the end-effector should be kept at the preset Euler angles $(\theta^p_x, \theta^p_y, \theta^p_z)$. Therefore, to maintain the synchronous motion of the robot end-effector and the moving target object in a relatively stationary state, the tracking action should include the tracking position $(x^c + x^p, y^c + y^p, z^c + z^p)$, tracking orientation $(\theta^p_x, \theta^p_y, \theta^p_z)$, and tracking velocity $(v_x, v_y, v_z)$.

By solely executing the tracking action as the manipulation action, the end-effector first approaches the target object, as visualized at Timestamp 1 of Fig.~\ref{Fig: hybrid_control}. Then, at Timestamp 2, the end-effector achieves a stable tracking status. In this status, the moving target object appears stationary in the view of the wrist camera. Upon attaining this stable status, as illustrated at Timestamp 3 of Fig.~\ref{Fig: hybrid_control}, the interaction action predicted by the GEM network begins to be sequentially added to the tracking action, which then serves as the composite manipulation action. The interaction action includes the position offset $(x^n, y^n, z^n)$ and pose offset $(\theta^n_x, \theta^n_y, \theta^n_z)$. In this control strategy, the tracking action alone ensures that the target object is in a relatively stationary state with respect to the robot end-effector. This means when the GEM network predicts interaction actions, it behaves as though it were manipulating a stationary object. In this way, the influence of various conveyor speeds on generating interaction actions is explicitly decoupled and removed. Therefore, this action decomposition strategy enables the GEM network trained using static object manipulation demonstrations to operate objects moving at diverse speeds on conveyor systems.

Summarily, before the end-effector achieves the stable tracking status, the manipulation action is identical to the tracking action computed from the point segmentation result, and it comprises the position goal $(x^c + x^p, y^c + y^p, z^c + z^p)$, orientation goal $(\theta^p_x, \theta^p_y, \theta^p_z)$, and velocity goal $(v_x, v_y, v_z)$. After stable tracking is attained, the manipulation action is the sum of the tracking action and interaction action, including the position goal $(x^n + x^c + x^p, y^n + y^c + y^p, z^n + z^c + z^p)$, orientation goal $(\theta^n_x + \theta^p_x, \theta^n_y + \theta^p_y, \theta^n_z + \theta^p_z)$, and velocity goal $(v_x, v_y, v_z)$. The manipulation action is transformed into the control signal of robot joints and end-effector based on inverse kinematics and PID \cite{bennett1993development}. Notably, the status flag predicted by the policy indicates whether the manipulation of an object has been completed. Once completed, the interaction action is removed from the manipulation action, and the tracking action switches to the next target object.

\subsection{Network Implementation}
\label{sub:network_implementation}

\begin{figure}[tbp]
    \centering
    \includegraphics[width=1.0\linewidth]{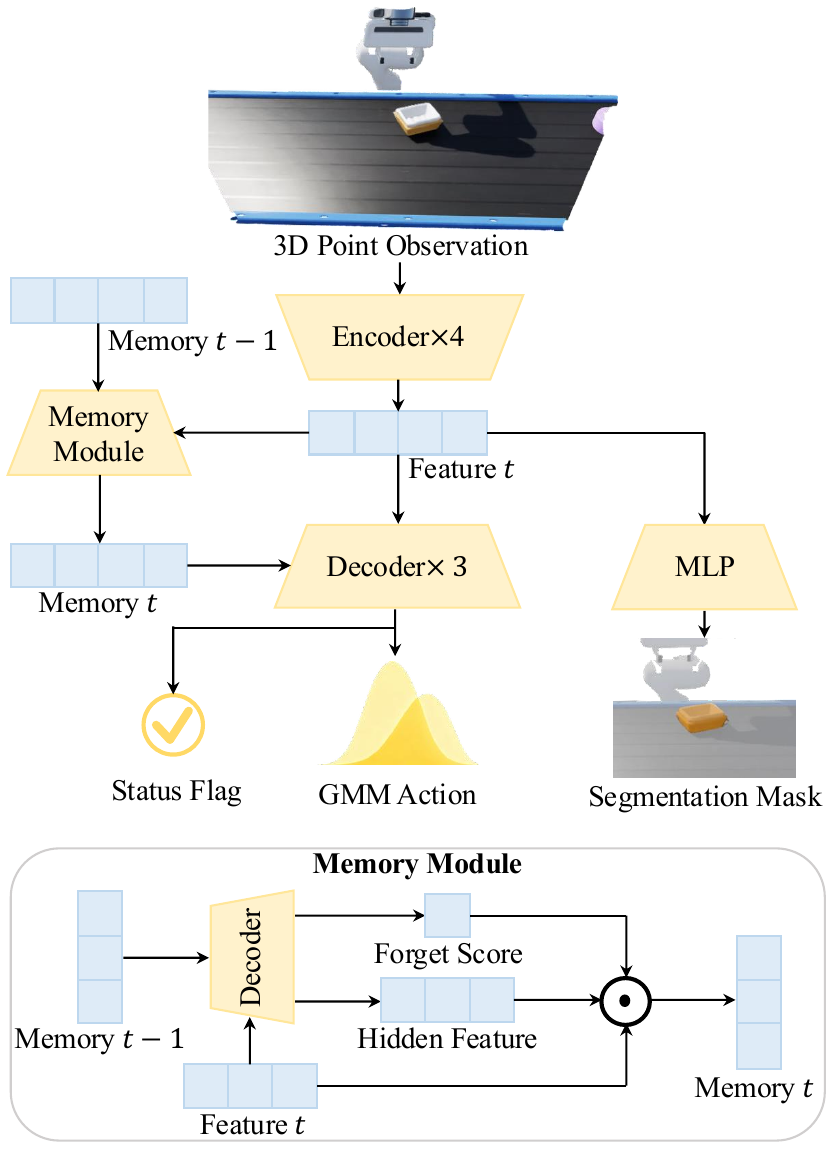}
    \caption{The overall structure of the GEM network, which consists of four point encoders, three Transformer decoders, and a memory module. The memory module updates the memory representation in a recurrent manner. The network outputs include the GMM-based action distribution, point segmentation mask, and a status flag that marks the completion of manipulating one object.} \label{Fig: network_pipeline}
\end{figure}

The overall framework of the GEM network is illustrated in Fig.~\ref{Fig: network_pipeline}. During training, the network input is the 3D points perturbed by the appearance noise annealing strategy. In inference, the input is the points without noise. For the 3D points perceived at timestamp $t$, the network encodes the points into a semantic-rich feature embedding $F_t \in \mathbb{R}^{l_p \times c}$ using four point encoders \cite{wu2024point}, where $l_p$ denotes the token number and $c$ is the token length. We initialize the parameters of these encoders with Sonata \cite{wu2025sonata}. Each token in $F_t$ corresponds to a voxel in the 3D physical space. In the point segmentation branch, a multilayer perceptron (MLP) layer is applied to transform $F_t$ into the segmentation mask by performing binary classification. The segmentation branch optimization objective is defined as a binary cross-entropy loss \cite{lecun2015deep}.

Subsequently, the memory module shown in Fig.~\ref{Fig: network_pipeline} fuses $F_t$ and the memory feature at timestamp $t-1$, denoted as $M_{t-1} \in \mathbb{R}^{l_m \times c}$, into the new memory feature $M_t \in \mathbb{R}^{l_m \times c}$. During this process, the memory encoder first employs a Transformer decoder \cite{carion2020end} to conduct feature interaction between $M_{t-1}$ and $F_t$. Specifically, $M_{t-1}$ serves as the query embedding, and $F_t$ is the key and value embeddings. The feature output by this Transformer decoder is transformed into the forget score $G_t \in \mathbb{R}^{l_m \times 1}$ and hidden feature $H_t \in \mathbb{R}^{l_m \times c}$ by two MLP layers. Afterwards, $M_t$ is calculated following:
\begin{align}
M_t = \sigma(G_t) \odot H_t + (1- \sigma(G_t)) \odot M_{t-1}, \label{Eq7}
\end{align}
where $\sigma(\cdot)$ represents the Sigmoid function. At the beginning of manipulation, the memory feature is initialized as a zero embedding. Incorporating historical memory into the network observation is beneficial for enabling GEM to accurately assess the current manipulation progress, thereby reducing the ambiguity in network optimization.

Then, $F_t$ and $M_t$ are concatenated and input to three Transformer decoders to conduct feature interaction with $N_a$ action tokens. The interaction process in decoders is similar to the implementation of ACT \cite{zhao2023learning}, where the $N_a$ action tokens correspond to the $N_a$ actions scheduled for the subsequent $N_a$ timestamps. After the interaction in these decoders, we transform these action tokens into the probabilistic distributions of actions. This operation is different from recent works that commonly decode action tokens into deterministic action values \cite{kimopenvla}. We choose to predict action distributions because this design better captures the multi-modal nature of embodied actions, thereby alleviating the optimization conflict in training caused by decision-making ambiguity.

For each action token, we decode it into an action distribution represented using GMM with $K$ components. Specifically, an MLP layer predicts the means, variances, and weights of these $K$ Gaussian components based on each action token. For the $t_{\rm th}$ ($t \in [1, N_a]$) action token, denote the mean, variance, and weight of the $k_{\rm th}$ component as $\mu_{t,k}$, $\sigma_{t,k}^2$, and $\alpha_{t,k}$, respectively. The distribution of the $t_{\rm th}$ scheduled action $p(a_t)$ can be formulated as:
\begin{align}
p(a_t) = \sum\limits_{k=1}^K \alpha_{t,k} \mathcal{N}(\mu_{t,k}, \sigma^2_{t,k}). \label{Eq8}
\end{align}
The component weights $\{\alpha_{t,k}\}_{k=1}^K$ satisfy $\sum\limits_{k=1}^K \alpha_{t,k} = 1$ and $\alpha_{t,k} \geq 0$ for all $k \in \{1, 2, \cdots, K\}$. In training, assuming the corresponding expert actions of the $N_a$ timestamps are $\{\tilde{a}_t\}_{t=1}^{N_a}$. According to Eq.~(\ref{Eq3}), the supervision loss for action prediction is defined as:
\begin{align}
L_a = -\frac{1}{N_a} \sum\limits_{t=1}^{N_a} \log (\sum\limits_{k=1}^{K} \frac{\alpha_{t,k}}{\sqrt{2 \pi \sigma_{t,k}^2}} \exp(-\frac{(\tilde{a}_t - \mu_{t, k})^2}{2 \sigma_{t, k}^2}) ). \label{Eq9}
\end{align}
During inference, the highest-probability action at each of the $N_a$ future timestamps is selected, forming an action chunk that is executed in a single rollout. Moreover, the GEM network predicts a Boolean status flag that indicates whether the manipulation of the current object is completed. As described in Section~\ref{sub:manipulation_action_decomposition}, this flag controls\textbf{} when the policy should switch to the next target object. In training, the status flag prediction is supervised with a binary cross-entropy loss.
 
\subsection{Data Generation in Simulation}
\label{sub:data_generation_in_simulation}

\begin{figure}[tbp]
    \centering
    \includegraphics[width=1.0\linewidth]{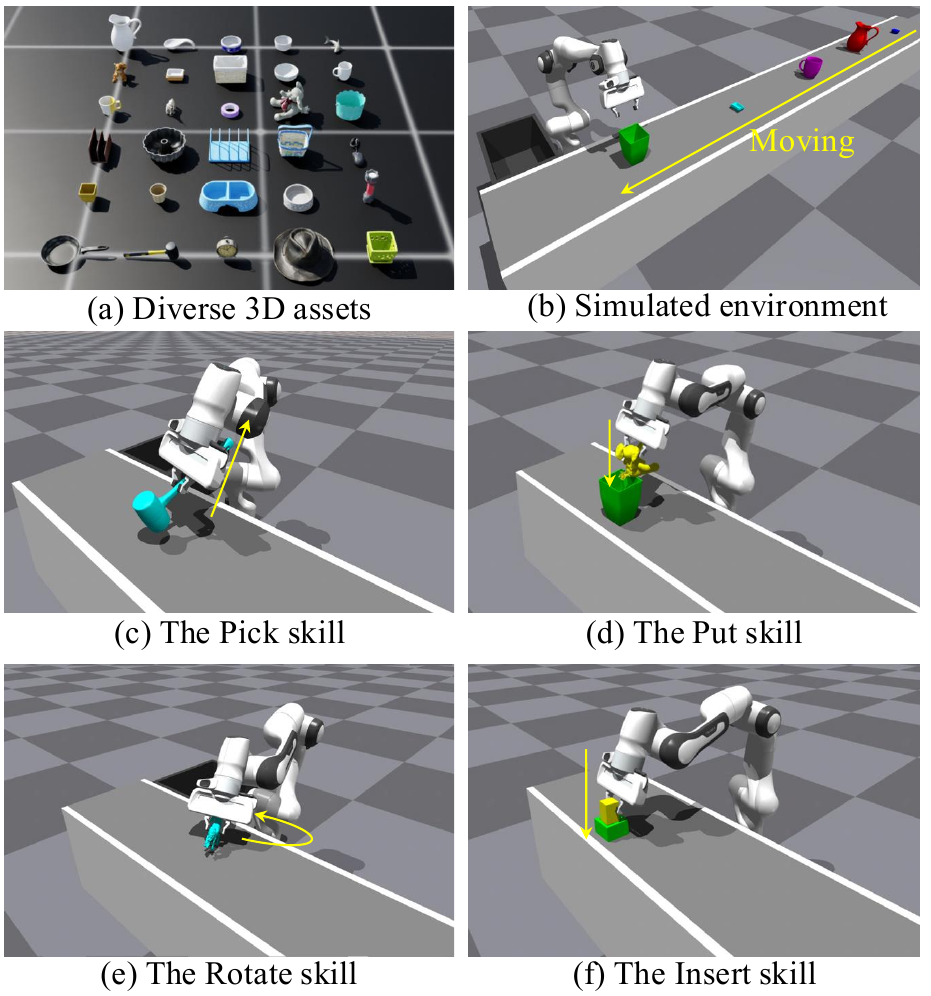}
    \caption{This figure illustrates the developed dynamic object manipulation simulator for generating demonstration data efficiently. Specifically, Subfigure~(a) shows the collected 3D assets of diverse geometries. Subfigure~(b) visualizes the simulated environment. Subfigure~(c)$\sim$(f) depict examples of the four supported manipulation skills, including Pick, Put, Rotate, and Insert.} \label{Fig: simulated_envi}
\end{figure}

Combining the aforementioned techniques, we derive an efficient policy that can realize sim-to-real generalization. In this subsection, we explain our built simulation environment for dynamic object manipulation on conveyor belts.

Thanks to the appearance noise annealing strategy, the GEM network primarily focuses on the geometric structure of objects, and its reliance on visual appearance is reduced. To ensure that the trained network can generalize to diverse unseen objects, we need to generate manipulation demonstrations using 3D assets of varied geometries. This is because a key advantage of our method is that while the texture appearances of objects are highly diverse and nearly impossible to exhaustively enumerate, the set of common geometries is relatively limited. Therefore, by collecting a set of 3D assets that covers common object geometries, our method can achieve promising generalization. This insight empowers us to avoid collecting a very large number of 3D assets, significantly reducing the manual effort required for 3D asset preparation. Based on this insight, we select 30 assets with representative geometries from the GSO dataset \cite{downs2022google} to generate demonstration data, and these assets are depicted in Fig.~\ref{Fig: simulated_envi}~(a).

We adopt Isaac Gym \cite{makoviychuk2isaac} to build the simulation environment for dynamic object manipulation, as visualized in Fig.~\ref{Fig: simulated_envi}~(b). We do not employ more visually photorealistic simulators like Isaac Sim \cite{nambiar2024automation} because we find their simulation fidelity on object interaction dynamics is inadequate for automatic data generation. Hence, in this work, Isaac Sim is mainly utilized to evaluate the out-of-domain generalization capabilities of policies trained using the demonstrations collected in Isaac Gym. Since Isaac Gym does not support rendering complex appearance textures, we assign random solid colors to objects during data collection, as illustrated in Fig.~\ref{Fig: simulated_envi}~(b). By contrast, as depicted in Fig.~\ref{Fig: simulated_envi}~(a), objects exhibit realistic textures during the evaluation in Isaac Sim.

Based on the developed simulation environment, we generate demonstrations of four common skills in dynamic object manipulation tasks on conveyor belts, including Pick, Put, Rotate, and Insert, which are shown in Fig.~\ref{Fig: simulated_envi}~(c)$\sim$(f), respectively. Specifically, Pick requires the robot to pick up the moving object from the conveyor belt. Place needs the robot to put an object into a moving container on the conveyor belt. Rotate requires the robot to rotate the moving object to a target orientation. The target orientation in this work is prompted to the network by providing a reference observation, which is the point cloud of the manipulated object in the goal 3D orientation from the perspective of the wrist camera. Insert involves inserting an object into a container with a precisely sized opening, ensuring the object is aligned at the specified pose. The Insert task illustrated in Fig.~\ref{Fig: simulated_envi}~(f) only allows two millimeters of position error. 

To enable automatic demonstration generation without human intervention, we employ CuRobo \cite{sundaralingam2023curobo} to generate manipulation trajectories based on the true states of objects obtained from the simulator. We collect 2,000 demonstrations for each of the four skills, and every demonstration comprises about ten seconds of manipulation actions, robot proprioception information, and environment observations.

\section{Experimental Verifications}
\label{sec:experimental_verifications}

\subsection{Experimental Settings}
\label{sub:experimental_settings}

\subsubsection{Implementation Details}
As described in Section~\ref{sub:data_generation_in_simulation}, we generate 2,000 demonstrations for each skill in Pick, Put, Rotate, and Insert. We obtain a total of 8,000 demonstrations for these four skills and utilize them to train networks. We optimize the GEM network for 10K optimization iterations. The training takes about 14 hours on two RTX4090 GPUs. The optimizer is AdamW \cite{loshchilovdecoupled} and the learning rate is $1e-4$. The training batch size is 32. The image resolutions of the two RGB-D cameras are both $640\times480$. The action chunk size $N_a$, memory token number $l_m$, and feature channel number $c$ are set to 20, 200, and 512, respectively. GEM controls the robot at a frequency of 20 HZ. Since objects are transported successively, this work does not consider scenarios where objects on the conveyor belt are cluttered together.

\begin{figure}[tbp]
    \centering
    \includegraphics[width=1.0\linewidth]{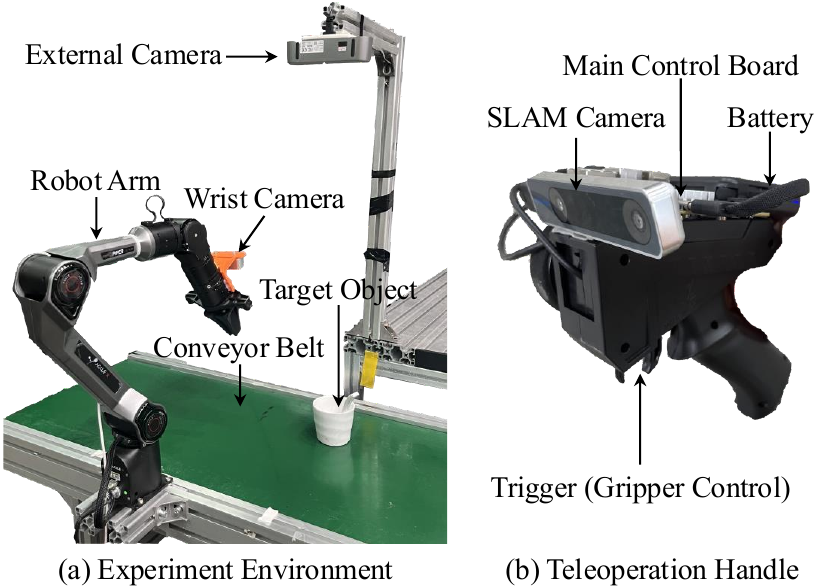}
    \caption{The real-robot experiment platform and remote teleoperation handle. Specifically, Sub-figure (a) illustrates the real-robot environment for studying dynamic object manipulation. GEM controls the robot arm to perform diverse tasks given the observations from the external and wrist RGB-D cameras. Sub-figure (b) shows our designed remote handle for teleoperation. In the experiments needing real-world tuning data, we utilize this remote handle to collect demonstrations efficiently. } \label{Fig: real_robot}
\end{figure}

\begin{figure}[tbp]
    \centering
    \includegraphics[width=1.0\linewidth]{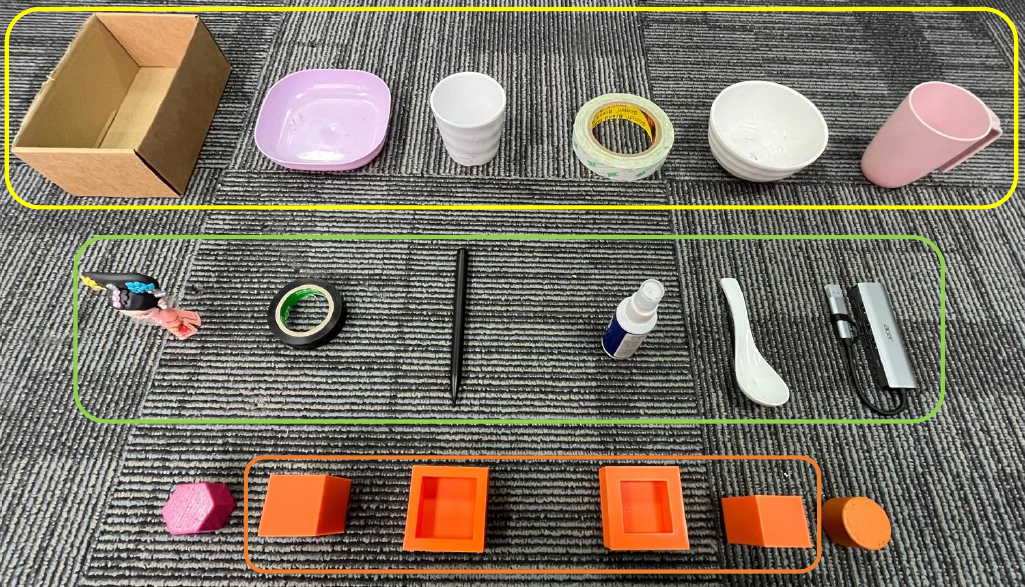}
    \caption{The 18 objects of diverse geometries selected to conduct real-world experiments. All the objects can be used in the evaluation of the skills Pick and Rotate. For Put, one of the small-sized objects highlighted in the green box is put in one of the containers in the yellow box. The objects in the orange box are for Insert, which only allows two millimeters of insertion error.} \label{Fig: real_assets}
\end{figure}

\begin{figure}[tbp]
    \centering
    \includegraphics[width=1.0\linewidth]{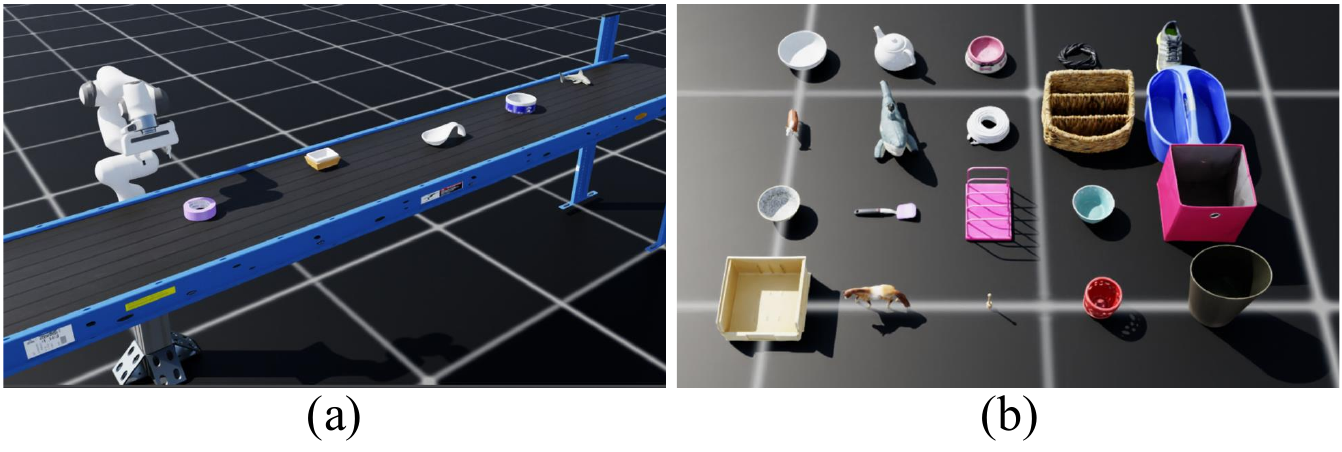}
    \caption{(a) We build a simulation environment in Isaac Sim to evaluate the out-of-domain generalization of policies trained using the demonstrations from Isaac Gym. (b) We select 20 additional assets to validate the manipulation performance of policies on unseen objects.} \label{Fig: IsaacSim_envi}
\end{figure}

\subsubsection{Real-Robot Platform}

To conduct real-robot experiments, we build a hardware platform as shown in Fig.~\ref{Fig: real_robot}. We employ this platform to evaluate the sim-to-real performance of policies trained using simulated demonstrations. In this platform, objects are driven by a conveyor belt at an adjustable velocity. A robot arm with a gripper is mounted on one side of this conveyor belt to manipulate the target object. Two RGB-D cameras are deployed to capture environment observations, one on the robot arm (the wrist camera) and the other on a fixed third-person perspective bracket (the external camera). GEM controls the robot arm to manipulate objects based on the observations from these two cameras.

As illustrated in Fig.~\ref{Fig: real_robot} (b), a teleoperation handle is designed to teleoperate the robot arm in Fig.~\ref{Fig: real_robot} (a). This handle is used in the experiments needing real-world tuning demonstrations. Specifically, we control the motion of the robot arm end-effector based on the relative pose change of the handle derived from the SLAM camera. The opening and closing of the gripper is regulated through the trigger pull degree. A battery is mounted on the handle to power all the devices in this handle. The control signals from the SLAM camera and trigger are wirelessly transmitted to the robot arm by the main control board, thereby enabling flexible remote teleoperation. The handle is not used when testing the method.

To evaluate the capabilities of GEM on manipulating objects in various geometries, we select 18 objects as shown in Fig.~\ref{Fig: real_assets} to conduct experiments. All the objects can be used to validate the skills Pick and Rotate. For Put, the robot needs to put one of the small-sized objects highlighted in the green box into one of the containers marked in the yellow box. The objects in the orange box are for Insert, which are in various sizes and only allow two millimeters of insertion error.

\subsubsection{Simulation Platform}

Although real-robot experiments directly reflect robot performance in the real world, it is inconvenient to adjust environment variables to conduct an ablation study. In contrast, simulation experiments allow controlling environmental variables precisely, so we also evaluate our method in simulation. As illustrated in Fig.~\ref{Fig: simulated_envi}~(a) and (b), we gather objects of diverse geometries and build a simulation environment in Isaac Gym to generate training demonstrations. We can also use this environment to validate the in-domain performances of trained policies. 

To further test the out-of-domain generalization results, we develop a simulation environment in Isaac Sim, as depicted in Fig.~\ref{Fig: IsaacSim_envi}~(a). We can find that the visual fidelity of the rendered environment in Isaac Sim is promising, ensuring the behaviors of the evaluated policies are similar to those in the real world. Notably, a problem with using Isaac Sim is the occasional occurrence of abnormal collision dynamics. For example, a gripper might lightly touch a cup, causing it to jump or move unrealistically. Due to this problem, we do not choose Isaac Sim to build the automatic demonstration generation pipeline. During evaluating policies in Isaac Sim, we manually inspect such simulation anomalies and exclude abnormal cases from the experimental statistics. In addition, as shown in Fig.~\ref{Fig: IsaacSim_envi} (b), we select 20 new objects to validate the manipulation success rates of policies on unseen objects.

\subsubsection{Testing Protocols} We analyze policies using both the aforementioned simulated environments and a real-world platform. The primary evaluation metric is the task completion success rate. In Pick, the objects in various geometric structures are successively conveyed on the conveyor belt. The success condition for Pick is the gripper successfully picks the moving object up. In Put and Insert, an object and a container are moving on a conveyor belt. Success is defined as the robot arm places the object inside the container. For Rotate, the robot hand grasps and rotates the object on the conveyor belt to the target orientation. Success is achieved if the yaw, pitch, and roll errors are all within 5 degrees. The default moving speed of the conveyor belt is 0.1 m/s. To mitigate the influence of random errors, we repeat each experiment for 100 times to compute the average success rate. 

In simulation, we validate policies based on the in-domain and out-of-domain settings. In the in-domain setting, the policies are evaluated in Isaac Gym to manipulate the objects that have appeared in training demonstrations. By contrast, for the out-of-domain setting, the policies manipulate the unseen objects shown as Fig.~\ref{Fig: IsaacSim_envi}~(b) in Isaac Sim environments. For real-robot experiments, we test policies with the tuned sim-to-real and zero-shot sim-to-real settings. In tuned sim-to-real, the policies pre-trained with simulated data are further fine-tuned with 10 real-world demonstrations before evaluation. Differently, the zero-shot sim-to-real setting requires the policies to operate a real robot without real-world data tuning.

\subsection{Method Comparison}
\label{sub:method_comparison}

This subsection compares GEM with previous manipulation methods in dynamic object manipulation. Although the classic kinematics-based methods can also handle dynamic object manipulation after precise system identification and careful parameter tuning, they demand special algorithm designs for different tasks and testing environments. This task-dependent characteristic violates the focus of this work, how to build a policy that is generalizable across various scenarios. Therefore, we mainly compare GEM with the recent learning-based methods, which present better generalization.

The compared methods include AnyGrasp \cite{fang2023anygrasp}, ConvMLP\cite{zhang2018deep}, RVT-2 \cite{goyal2024rvt}, ACT \cite{zhao2023learning}, Diffusion Policy (DP) \cite{chi2023diffusion}, Diffusion Policy 3D (DP3) \cite{ze20243d}, OpenVLA \cite{kimopenvla}, and PI0 \cite{black2024pi_0}. Among them, AnyGrasp can generate grasp pose candidates based on the 3D points of the target object. Then, the robot arm selects the grasp pose with the highest confidence and grasps the target based on kinematic control. Since AnyGrasp only predicts the grasp pose of each object once, the grasp is executed in an open-loop manner. RVT-2 is a representative two-stage method that takes RGB-D information as input and processes this information based on point cloud reconstruction. Similar to the GEM network, DP3 takes 3D points as environmental observation, but it does not employ visual color feature. The inputs of the other compared methods are the wrist and external view images. Among them, ConvMLP, ACT, and DP are lightweight neural networks based on image observations. OpenVLA and PI0 are two influential vision-language-action (VLA) models. As OpenVLA and PI0 contain billions of parameters, their inference speeds are limited.

\begin{table*}[tbp] 
    \centering
    \caption{Success rate and speed comparison with previous methods in simulation.} \label{Table: Simulation Comparison}
    \vspace{-0.2cm}
    \tabcolsep=0.15cm
    \resizebox{0.9\linewidth}{!}{
    \begin{tabular}{c|ccccc|ccccc|cc}
    \toprule
    Method & Pick$\uparrow$ & Put$\uparrow$ & Rotate$\uparrow$ & Insert$\uparrow$ & Avg\_in$\uparrow$ & Pick\_o$\uparrow$ & Put\_o$\uparrow$ & Rotate\_o$\uparrow$ & Insert\_o$\uparrow$ & Avg\_out$\uparrow$ & Avg$\uparrow$ & Speed (HZ)$\uparrow$ \\
    \midrule
    AnyGrasp* & 0.41 & - & - & - & - & 0.46 & - & - & - & - & - & 4.35 \\
    ConvMLP & 0.06 & 0.19 & 0.05 & 0.00 & 0.08 & 0.00 & 0.03 & 0.00 & 0.00 & 0.01 & 0.05 & 17.54 \\
    ConvMLP* & 0.28 & 0.48 & 0.19 & 0.01 & 0.24 & 0.03 & 0.24 & 0.02 & 0.00 & 0.08 & 0.16 & 13.16 \\
    RVT-2 & 0.08 & 0.24 & 0.03 & 0.00 & 0.09 & 0.00 & 0.07 & 0.00 & 0.00 & 0.02 & 0.05 & 31.48 \\
    RVT-2* & 0.47 & 0.62 & 0.15 & 0.27 & 0.38 & 0.12 & 0.40 & 0.00 & 0.05 & 0.14 & 0.26 & 19.69 \\
    ACT & 0.14 & 0.29 & 0.10 & 0.00 & 0.13 & 0.02 & 0.19 & 0.00 & 0.00 & 0.05 & 0.09 & 43.48 \\
    ACT* & 0.62 & 0.75 & 0.53 & 0.44 & 0.59 & 0.20 & 0.39 & 0.07 & 0.23 & 0.22 & 0.41 & 23.70 \\
    DP & 0.09 & 0.24 & 0.07 & 0.00 & 0.10 & 0.00 & 0.15 & 0.00 & 0.00 & 0.04 & 0.07 & 27.32 \\
    DP* &  0.55 & 0.70 & 0.46 & 0.36 & 0.52 & 0.17 & 0.32 & 0.03 & 0.18 & 0.18 & 0.35 & 17.86 \\
    DP3 & 0.31 & 0.50 & 0.29 & 0.00 & 0.28 & 0.19 & 0.40 & 0.22 & 0.00 & 0.20 & 0.24 & 29.40 \\
    DP3* & 0.72 & 0.85 & 0.77 & 0.70 & 0.76 & 0.57 & 0.74 & 0.62 & 0.44 & 0.59 & 0.68 & 18.87 \\
    OpenVLA & 0.11 & 0.26 & 0.08 & 0.00 & 0.11 & 0.06 & 0.18 & 0.01 & 0.00 & 0.06 & 0.09 & 1.61 \\
    OpenVLA* & 0.64 & 0.74 & 0.46 & 0.35 & 0.55 & 0.31 & 0.64 & 0.05 & 0.24 & 0.31 & 0.43 & 1.58 \\
    PI0 & 0.23 & 0.36 & 0.18 & 0.00 & 0.19 & 0.09 & 0.24 & 0.06 & 0.00 & 0.10 & 0.15 & 1.97 \\
    PI0* & 0.68 & 0.81 & 0.59 & 0.65 & 0.68 & 0.47 & 0.70 & 0.11 & 0.38 & 0.42 & 0.55 & 1.92 \\
    GEM (Ours) & 0.94 & 1.00 & 0.90 & 0.86 & 0.93 & 0.86 & 0.96 & 0.90 & 0.83 & 0.89 & 0.91 & 23.81 \\
    \bottomrule
    \end{tabular}}
\end{table*}

\begin{table}[tbp] 
    \centering
    \caption{Success rate comparison with previous methods in real world tasks.} \label{Table: Real Comparison}
    \vspace{-0.2cm}
    \tabcolsep=0.25cm
    \resizebox{0.9\linewidth}{!}{
    \begin{tabular}{c|ccccc}
    \toprule
    Method & Pick$\uparrow$ & Put$\uparrow$ & Rotate$\uparrow$ & Insert$\uparrow$ & Avg$\uparrow$ \\
    \midrule
    AnyGrasp* & 0.45 & - & - & - & - \\
    ConvMLP & 0.08 & 0.25 & 0.07 & 0.00 & 0.10 \\
    ConvMLP* & 0.36 & 0.55 & 0.28 & 0.13 & 0.33 \\
    RVT-2 & 0.13 & 0.30 & 0.09 & 0.00 & 0.13 \\
    RVT-2* & 0.51 & 0.68 & 0.41 & 0.28 & 0.47 \\
    ACT & 0.18 & 0.34 & 0.20 & 0.00 & 0.18 \\
    ACT* & 0.65 & 0.76 & 0.55 & 0.49 & 0.61 \\
    DP & 0.15 & 0.29 & 0.13 & 0.00 & 0.14 \\
    DP* & 0.61 & 0.69 & 0.56 & 0.32 & 0.55 \\
    DP3 & 0.36 & 0.45 & 0.39 & 0.04 & 0.31 \\
    DP3* & 0.70 & 0.80 & 0.72 & 0.61 & 0.71 \\
    OpenVLA & 0.00 & 0.02 & 0.05 & 0.00 & 0.02 \\
    OpenVLA* & 0.32 & 0.39 & 0.20 & 0.03 & 0.24 \\
    PI0 & 0.03 & 0.08 & 0.14 & 0.00 & 0.06 \\
    PI0* & 0.40 & 0.48 & 0.56 & 0.11 & 0.39 \\
    GEM (Ours) & 0.98 & 1.00 & 0.93 & 0.89 & 0.95 \\
    \bottomrule
    \end{tabular}}
\end{table}

\subsubsection{Simulated Comparison}
This part compares GEM with the aforementioned counterparts in simulation. Since AnyGrasp is for generating grasp poses, we only test it in Pick tasks. The other compared methods are trained using the same simulated demonstrations as our GEM network. Since the compared models have no designs for dynamic object manipulation, we test them in two modes, the sole mode and hybrid mode. The sole mode means the robot movement is solely controlled by their network outputs. In the hybrid mode, we utilize a segmentation model \cite{khanam2024yolov11} to estimate the spatial position of the target object and then employ a visual servo control algorithm \cite{papanikolopoulos1993visual} that controls the end-effector to track the target object. The control signal is the sum of the network output and the visual servo control module, which shares a similar process as our proposed manipulation action decomposition strategy. In this way, the original dynamic object manipulation problem is converted into a static object manipulation problem, significantly improving the performance of the compared methods. As AnyGrasp is an open-loop method, we only report its results in the hybrid mode.

As described in Section~\ref{sub:experimental_settings}, we compare these policies in the in-domain setting (manipulate seen objects in Isaac Gym) and out-of-domain setting (manipulate unseen objects in Isaac Sim). The success rates and inference speeds of all the methods are presented in Table~\ref{Table: Simulation Comparison}. Specifically, the $2_{\rm nd}-6_{\rm th}$ columns are the success rates of the in-domain setting. The $7_{\rm th}-11_{\rm th}$ columns correspond to the out-of-domain setting. Avg\_in, Avg\_out, and Avg refer to the average success rates of in-domain tests, out-of-domain tests, and all tests, respectively. The inference speeds reported in the $13_{\rm th}$ column are the model inference frequencies tested on an RTX4090 GPU. An asterisk (*) following a method name indicates that the method is validated in the hybrid mode. The hybrid mode increases the inference time compared to the sole mode (denoted without the asterisk) because of the additional adoption of an instance segmentation model.

According to the results, we can find that GEM achieves the best results in performing all the skills. This is partly because of the design that helps the GEM network prioritize geometry information. In addition, it is observed that the methods tested in the hybrid mode significantly outperform those of the sole mode, which suggests the effectiveness of our action decomposition strategy. For AnyGrasp, we find it generates promising pick poses. However, it works in an open-loop manner, which restricts its real-time response to object pose changes. Though RVT-2 uses depth information like our method, its network architecture is inefficient in geometry information exploitation, resulting in its unsatisfactory success rates. Comparing the results of the VLA models and other lightweight networks like ACT, we observe that VLAs generally present better out-of-domain generalization because they are pre-trained on numerous data. However, the speeds of VLAs are slower. The slow inference does not affect the success rates significantly in simulated environments because the simulation is paused during model inference, but it harms performance in real-world tasks significantly. Compared with the other policies, the success rates of DP3 are promising although its structure is simple. This is because it takes 3D points without visual texture as input, and this strategy also prioritizes 3D geometry.

\subsubsection{Real Robot Comparison}
This study compares GEM with the aforementioned counterparts in real-robot tasks using the platform shown in Fig.~\ref{Fig: real_robot}~(a). We only report the results in the tuned sim-to-real setting, which employs 10 real-world demonstrations to fine-tune policies before evaluation. This is because we find the compared methods mostly cannot generalize to real-world manipulation after training with simulated data, though GEM generalizes well since it prioritizes geometry information. The tuning step adapts the policy to the real-world embodiment and environment observations.

Similar to the simulation experiments, we validate the compared methods in both the sole and hybrid modes. The real robot comparison results are presented in Table~\ref{Table: Real Comparison}, where we can find that GEM presents high success rates. According to the results, the sim-to-real performance of GEM outperforms the compared counterparts by large margins. Besides, the results of the compared methods in the hybrid mode significantly outperform the results in the sole mode, indicating the value of our manipulation action decomposition. Moreover, we can find that although PI0 behaves better than ACT in simulated environments, its success rates are lower in real robot experiments. This is caused by the slow inference of PI0. Therefore, fast inference speed is important for models used in dynamic object manipulation.

\begin{table}[tbp] 
    \centering
    \caption{Analysis on the appearance noise annealing} \label{Table: appearance noise annealing}
    \vspace{-0.1cm}
    \tabcolsep=0.25cm
    \resizebox{0.95\linewidth}{!}{
    \begin{tabular}{c|cccc}
    \toprule
    Policy & Sim\_in$\uparrow$ & Sim\_out$\uparrow$ & Real\_tune$\uparrow$ & Real\_zero$\uparrow$ \\
    \midrule
    GEM-N & 0.91 (0.87) & 0.70 (0.24) & 0.86 (0.77) & 0.65 (0.16) \\
    GEM-S & 0.86 (0.69) & 0.84 (0.63) & 0.91 (0.74) & 0.83 (0.60) \\
    GEM-O & 0.93 (0.86) & 0.89 (0.83) & 0.95 (0.89) & 0.87 (0.72) \\
    \bottomrule
    \end{tabular}}
\end{table}

\subsection{Component Analysis}
\label{sub:component_analysis}

This subsection ablates the designs in GEM. When reporting the results, the four evaluation settings described in Section~\ref{sub:experimental_settings}, including simulated in-domain, simulated out-of-domain, real-world tuned sim-to-real, and real-world zero-shot sim-to-real, are denoted as Sim\_in, Sim\_out, Real\_tune, and Real\_zero, respectively. Due to the space limit, when reporting the manipulation success rates of these four settings, we report the average success rates of all skills.

\subsubsection{Analysis on Appearance Noise Annealing}

This part analyzes how the appearance noise annealing strategy improves the sim-to-real generalization of GEM. As described in Section~\ref{sub:appearance_noise_annealing}, we apply strong color perturbation to 3D points at the beginning of training and gradually remove the noise as training progresses. To study this strategy, we compare its performance against two baselines, one trained without any color perturbation and another trained with strong color perturbation applied throughout the entire training process. The experimental results are presented in Table~\ref{Table: appearance noise annealing}. The original GEM implementation, the GEM without color perturbation, and the GEM with strong color perturbation all the time are denoted as GEM-O, GEM-N, and GEM-S, respectively. The first number in every table cell is the average success rate of four skills, and the second number in a bracket represents the success rate of the Insert skill. We separately list the success rates of the Insert skill because this skill requires highly precise operation and the numerical result change is obvious.

Comparing GEM-O and GEM-N, we can observe that the appearance noise annealing strategy boosts the generalization of the GEM network in unseen environments significantly. For example, in the zero-shot sim-to-real setting of real-robot experiments, the average success rate over different skills is improved by 22\%. According to the results of GEM-O and GEM-S, gradually removing the color perturbation achieves better performance than constantly applying strong color perturbation. This is because the visual appearance information is useful in scenarios like distinguishing two closely placed objects. For example, in the Insert task, the network needs to discriminate the inserted object and container, and solely relying on object geometry is difficult to meet this requirement. Therefore, we can find from Table~\ref{Table: appearance noise annealing} that the Pick success rates of GEM-S are significantly weaker than GEM-O.

\begin{table}[tbp] 
    \centering
    \caption{Comparison between point and image observations.} \label{Table: input comparison}
    \vspace{-0.1cm}
    \tabcolsep=0.25cm
    \resizebox{0.9\linewidth}{!}{
    \begin{tabular}{c|cccc}
    \toprule
    Input & Sim\_in$\uparrow$ & Sim\_out$\uparrow$ & Real\_tune$\uparrow$ & Real\_zero$\uparrow$ \\
    \midrule
    Image & 0.71 & 0.35 & 0.41 & 0.39 \\
    Point & 0.93 & 0.89 & 0.95 & 0.87 \\
    \bottomrule
    \end{tabular}}
\end{table}

\subsubsection{Analysis on Point based Representation}

According to Table~\ref{Table: appearance noise annealing}, the GEM policy can generalize from simulated data to real-world manipulation to some extent even without the appearance noise annealing strategy. This characteristic is attributed to the 3D point observation of GEM rather than using images. As analyzed in Section~\ref{sub:optimization_path_shaping}, 3D points directly deliver geometry, which is inherently of higher predictivity for estimating manipulation actions. Therefore, by utilizing 3D point observation, GEM naturally prioritizes geometry information over visual appearance, and our developed appearance noise annealing strategy strengthens this prioritization. By contrast, if taking images as input, the network has to infer 3D geometry from visual appearance, meaning that the network must rely on exploiting visual appearance features. 

In this experiment, we study the performance gain brought by adopting point observation compared with employing image observation. When the network input is of the image modality, the GEM network encoders are switched to DINOv2 \cite{oquab2024dinov2}, which is a popular image backbone with discriminative representation. The experimental results of adopting point and image observations are presented in Table~\ref{Table: input comparison}.

According to the results, we can find that the models taking 3D points as input outperform the ones with images significantly. For example, in the zero-shot sim-to-real setting, the point-based policy surpasses the image-based policy by 48\%, which confirms the importance of using 3D point observation.

\begin{figure}[tbp]
    \centering
    \includegraphics[width=1.0\linewidth]{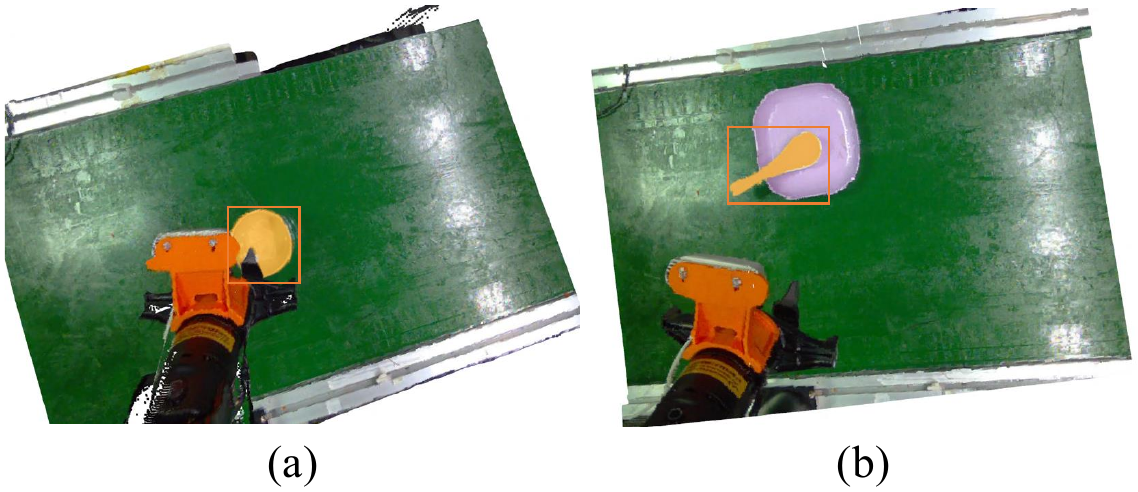}
    \caption{This figure visualizes some instance segmentation cases during manipulation. We render the segmented points in orange and highlight the segmentation mask with an additional orange box.} \label{Fig: inst_seg}
\end{figure}

\subsubsection{Analysis on Object Segmentation}

As depicted in Fig.~\ref{Fig: network_pipeline}, the GEM network predicts the 3D point segmentation mask of the target object to manipulate, and this segmentation mask is employed to generate tracking actions. Therefore, whether the network can produce a coarse mask of the target object is important for stably tracking the target object. 

To enable this stable tracking, we need promising segmentation mask annotations in training data. Thankfully, since we employ simulators to produce demonstrations, we can obtain precise segmentation labels automatically. Besides, since the GEM network primarily utilizes geometry feature to identify the target object, we find the network trained solely using simulation data generalizes to the real world well. To validate this, we collect a real-world segmentation dataset consisting of 1800 point scenes based on the assets shown in Fig.~\ref{Fig: real_assets}. Then, we utilize SAM2 \cite{ravi2024sam}, a state-of-the-art open-world segmentation model, to generate segmentation labels efficiently. The obtained labels are employed to validate the GEM network. Following the testing protocol of the COCO benchmark \cite{lin2014microsoft}, the GEM network achieves a Mask AP of 86.5\%, which meets the requirement of stable tracking.

Moreover, we visualize several target object segmentation examples in Fig.~\ref{Fig: inst_seg}, where the segmentation mask is highlighted in orange. The results demonstrate promising segmentation performance. As shown in Fig.~\ref{Fig: inst_seg}~(a), in the Pick or Rotate task that the operation involves only a single object, the network accurately segments that object. In Fig.~\ref{Fig: inst_seg}~(b), for the Put or Insert task that a gripper places an item into a container, the network successfully identifies the item based on its geometry characteristics.

\begin{table}[tbp] 
    \centering
    \caption{Analysis on the effect of the memory module.} \label{Table: memory module}
    \vspace{-0.1cm}
    \tabcolsep=0.2cm
    \resizebox{0.9\linewidth}{!}{
    \begin{tabular}{c|cccc}
    \toprule
    Memory & Sim\_in$\uparrow$ & Sim\_out$\uparrow$ & Real\_tune$\uparrow$ & Real\_zero$\uparrow$ \\
    \midrule
    $\times$ & 0.86 & 0.76 & 0.90 & 0.82 \\
    $\checkmark$ & 0.93 & 0.89 & 0.95 & 0.87 \\
    \bottomrule
    \end{tabular}}
\end{table}

\subsubsection{Analysis on Memory Modeling}
In dynamic object manipulation, modeling historical memory enhances the policy’s awareness of both the object’s motion dynamics and task progress. Therefore, we devise the memory module shown in Fig.~\ref{Fig: network_pipeline}. In experiments, we find that without this module, the robot arm could oscillate at the initial position of manipulation without approaching the target, leading to task failure. This phenomenon is more frequent when a single network learns multiple skills. This is because the policy fails to distinguish whether the manipulation operation has not started or has been completed. Our designed memory module can effectively address this problem by enabling the network to know which actions have been performed. We compare the performances of GEM with and without this memory module and report the results in Table~\ref{Table: memory module}. According to the results, we can observe that the memory module significantly boosts success rates.

To more thoroughly analyze how long a sequence the memory module can memorize, we design a toy experiment of reciting a random digit sequence. Specifically, we update the structure of this memory module to support accepting a single digit as input and predicting the next digit per step. Then, we randomly generate a sequence of $n$ digits and train the memory module to recite this sequence. After training, we test how many digits can be correctly recited. After repeating this experiment by setting different values of $n$, we find that the memory module can correctly recite a very long sequence like 1000 digits after sufficient epochs of training. Such a memory capability is strong enough for existing embodied policies, because current manipulation tasks usually involve at most tens of times of policy inference.

\begin{figure}[tbp]
    \centering
    \includegraphics[width=0.7\linewidth]{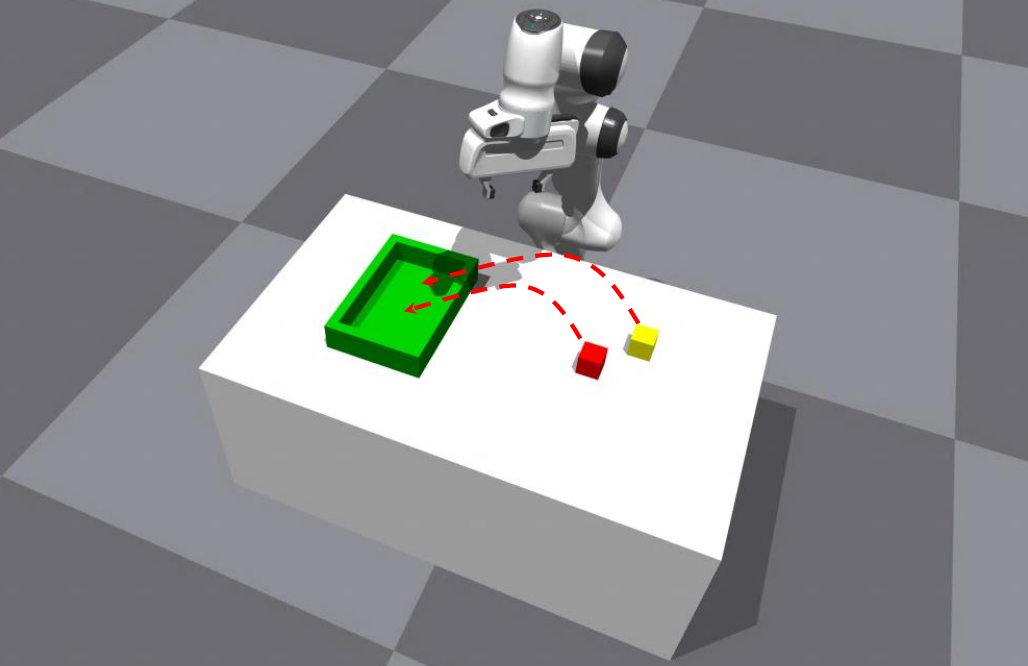}
    \caption{This figure visualizes the simulated environment for analyzing manipulation multimodal ambiguity. During collecting demonstrations in this environment, we teleoperate the robot hand to randomly pick up one of these two blocks and put it in the green container.} \label{Fig: two_box_envi}
\end{figure}

\begin{table}[t] 
    \centering
    \caption{Analysis on network output.} \label{Table: network output}
    \vspace{-0.1cm}
    \tabcolsep=0.2cm
    \resizebox{0.9\linewidth}{!}{
    \begin{tabular}{c|cccc}
    \toprule
    Output & Sim\_in$\uparrow$ & Sim\_out$\uparrow$ & Real\_tune$\uparrow$ & Real\_zero$\uparrow$ \\
    \midrule
    Direct & 0.88 & 0.80 & 0.91 & 0.79 \\
    DP & 0.81 & 0.76 & 0.85 & 0.63 \\
    FM & 0.86 & 0.77 & 0.88 &  0.75 \\
    Probabilistic & 0.93 & 0.89 & 0.95 & 0.87 \\
    \bottomrule
    \end{tabular}}
\end{table}

\subsubsection{Analysis on Network Output}
As explained in Section~\ref{sub:network_implementation}, the output of the GEM network is a GMM-based probabilistic distribution to capture the multi-modal nature of actions. This is different from previous works that directly regress the actions to execute \cite{fu2024mobile,chi2023diffusion}. This part analyzes the benefits of adopting this GMM-based probabilistic output.

We first compare the performances of the GEM network outputs based on direct action prediction, diffusion process (DP), flow matching (FM), and GMM. The results are reported in Table~\ref{Table: network output}. We can observe that the GMM-based network outperforms the other counterparts employing the compared output strategies, because the GMM output effectively alleviates the multimodal ambiguity problem. In addition, it is found that the performances of the DP and FM based outputs are inferior to direct action prediction, though DP and FM can also handle the multimodal ambiguity problem. After analysis, we infer this is because the denoising process in DP and FM is challenging for optimization, so a large training batch size is demanded to achieve promising manipulation action generation. In this experiment, since the training demonstration volume is limited, we cannot set such a large batch size. Thus, the obtained success rates are limited. According to these findings, we can summarize that the requirement of a large training batch size makes DP and FM less flexible than our developed GMM-based network output.

Notably, the results in Table~\ref{Table: network output} do not fully demonstrate the advantage of probabilistic action prediction. This is because the objects to manipulate arrive sequentially in dynamic object manipulation on conveyor belts. Hence, at a single moment, the policy usually observes only one object that is suitable for manipulation. As a result, the multi-modal ambiguity that the probabilistic head is designed to resolve remains limited.

To more sufficiently reveal the value of probabilistic action prediction, we develop a simulated environment as shown in Fig.~\ref{Fig: two_box_envi}. We use the teleoperation handle shown in Fig.~\ref{Fig: real_robot}~(b) to control the robot hand to collect 50 demonstrations, where we randomly pick up one of these two blocks and put it into the green container. After training using the collected demonstrations, we compare the success rates of the networks based on direct action prediction and probabilistic action prediction. We find that the one based on direct action prediction makes the robot end-effector hover at the midpoint between the two blocks, leading to a success rate of nearly 0\%. By contrast, the network based on probabilistic action prediction achieves a success rate of 87\%. These results demonstrate the importance of adopting our proposed probabilistic action prediction.

\begin{table}[tbp] 
    \centering
    \caption{Analysis on hybrid action control.} \label{Table: hybrid control}
    \vspace{-0.1cm}
    \tabcolsep=0.2cm
    \resizebox{0.95\linewidth}{!}{
    \begin{tabular}{c|cccc}
    \toprule
    Decomposition & Sim\_in$\uparrow$ & Sim\_out$\uparrow$ & Real\_tune$\uparrow$ & Real\_zero$\uparrow$ \\
    \midrule
    $\times$ & 0.37 & 0.31 & 0.45 & 0.30 \\
    $\checkmark$ & 0.93 & 0.89 & 0.95 & 0.87 \\
    \bottomrule
    \end{tabular}}
\end{table}

\subsubsection{Analysis on Manipulation Action Decomposition}
In GEM, we design a strategy that decomposes the manipulation actions into interaction actions and tracking actions, where the interaction actions are decoded from the predicted GMM distribution and the tracking actions are produced based on the point segmentation result. Within this strategy, the tracking actions control the robot end-effector to follow the motion of the target object. In this way, the pose of the end-effector keeps stationary relative to the target object, and thus the interaction actions can be regarded as operating a static object. This design significantly simplifies this task.

The experimental results in Table~\ref{Table: Simulation Comparison} and Table~\ref{Table: Real Comparison} indicate the effectiveness of this strategy. In these two tables, the compared policies marked with an asterisk (*) employ the action decomposition strategy, whereas the ones without an asterisk do not. We can find that the developed strategy improves their success rates by large margins. In addition, we also present the results of GEM without and with the manipulation action decomposition strategy in Table~\ref{Table: hybrid control}. According to the results, it is observed that this strategy is important for the overall performance of GEM.

\begin{figure}[tbp]
    \centering
    \includegraphics[width=1.0\linewidth]{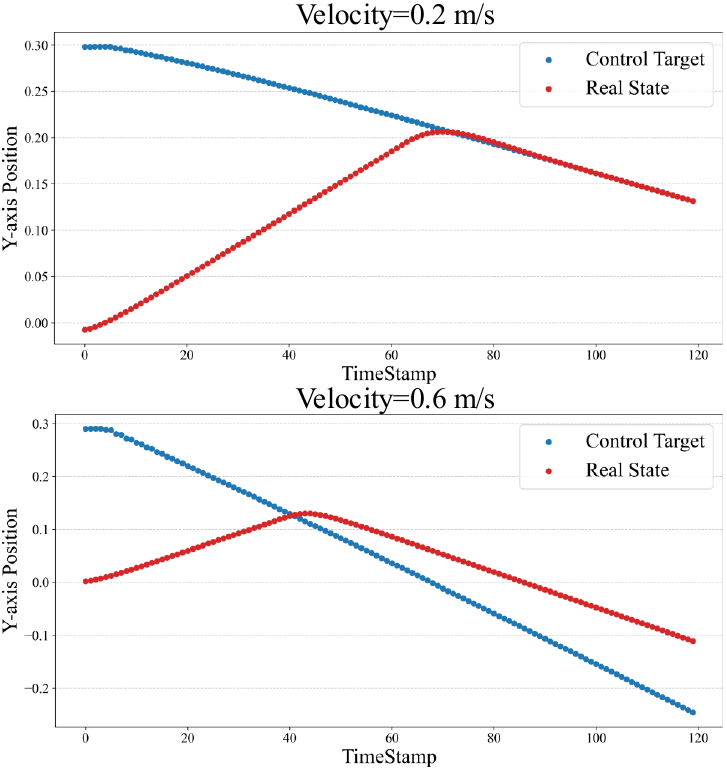}
    \caption{This figure visualizes the tracking dynamics of the end-effector under different target velocities. We can find that the end-effector stably tracks objects moving at 0.2 m/s but fails to keep up with objects moving at 0.6 m/s.} \label{Fig: track_control}
\end{figure}

\begin{table}[tbp] 
    \centering
    \caption{Manipulation performance in various moving speeds.} \label{Table: speed analysis}
    \vspace{-0.2cm}
    \tabcolsep=0.25cm
    \resizebox{0.8\linewidth}{!}{
    \begin{tabular}{c|cccc}
    \toprule
    Speed (m/s) & Pick$\uparrow$ & Put$\uparrow$ & Rotate$\uparrow$ & Insert$\uparrow$ \\
    \midrule
    0.05 & 0.92 & 1.00 & 0.91 & 0.87 \\
    0.10 & 0.94 & 1.00 & 0.90 & 0.86 \\
    0.25 & 0.86 & 0.98 & 0.85 & 0.63 \\
    0.50 & 0.00 & 0.00 & 0.00 & 0.00 \\
    \bottomrule
    \end{tabular}}
\end{table}

\subsubsection{Analysis on Object Tracking}
This work studies dynamic object manipulation on conveyor systems, where the target objects could move at various velocities. As discussed in the above paragraphs, we design a strategy that decomposes the manipulation actions into interaction actions and tracking actions to handle the diverse motion dynamics of target objects. In this part, we validate the performance of GEM on manipulating objects of different velocities.

First of all, we analyze the tracking stability and velocity upper limit in real-world manipulation. Specifically, we adjust the conveyor belt to various speeds and make the robot hand track a moving object based on the visual servo tracking module. It is observed that the maximum stable tracking speed in simulation is approximately 0.25 m/s, and it reaches about 0.3 m/s in the real world. In Fig.~\ref{Fig: track_control}, we visualize the dynamic curves of the control target signal and corresponding end-effector position. We can find that the end-effector approaches the moving object fast and tracks it stably under the velocity of 0.2 m/s. However, the end-effector fails to track the object when the velocity is 0.6 m/s due to its speed limit.

Subsequently, we validate the success rates of GEM on manipulating objects of different moving speeds in the simulated in-domain experiment setting. The results are reported in Table~\ref{Table: speed analysis}. Comparing the $1_{\rm st}$ and $2_{\rm nd}$ rows of results, we can observe that the success rates are similar when the conveyor belt runs at a low speed. When the speed is 0.25 m/s, the success rates drop because the robot velocity limit prevents the end-effector from maintaining stable tracking while simultaneously performing precise manipulation. When the speed arrives at 0.50 m/s, the robot hand cannot perform any manipulation task because it cannot keep pace with the movement of the target object. 

\subsection{Generalization Analysis}
\label{sub:generalization_analysis}

Thanks to the many designs in GEM, it shows strong generalization in various unseen scenarios, making it suitable for real applications. This subsection analyzes the generalization capabilities of GEM from different perspectives.

\begin{figure}[tbp]
    \centering
    \includegraphics[width=1.0\linewidth]{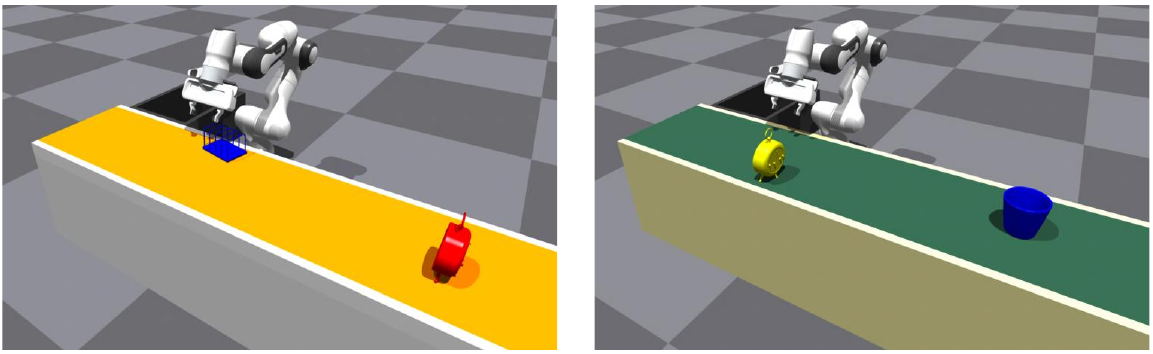}
    \caption{To study the background generalization of GEM while excluding the influence of irrelevant factors, we randomly augment the appearance of the conveyor belt in Isaac Gym. The augmented appearances are unseen in demonstrations. Two augmented samples are visualized in this figure.} \label{Fig: diverse_backgrounds}
\end{figure}

\begin{table}[tbp] 
    \centering
    \caption{Environment generalization study in simulation.} \label{Table: environment generalization study}
    \vspace{-0.2cm}
    \tabcolsep=0.25cm
    \resizebox{0.8\linewidth}{!}{
    \begin{tabular}{c|cccc}
    \toprule
    Environment & Pick$\uparrow$ & Put$\uparrow$ & Rotate$\uparrow$ & Insert$\uparrow$ \\
    \midrule
    Seen & 0.94 & 1.00 & 0.90 & 0.86 \\
    Unseen & 0.95 & 1.00 & 0.88 & 0.85 \\
    \bottomrule
    \end{tabular}}
\end{table}

\subsubsection{Generalization across Environment Backgrounds}

In applications, various scenarios show different environment backgrounds. It is expensive to collect demonstrations in every scenario to tune a policy. Thus, we expect that a policy can generalize to unseen scenarios for supporting more flexible deployment. Considering this expectation, we develop GEM, which prioritizes geometry structure and decreases the importance of visual appearance. In this part, we analyze the generalization capability of GEM in unseen backgrounds. 

Although we develop the out-of-domain simulation testing protocol, where the policy is trained using demonstrations from Isaac Gym and tested in Isaac Sim, this protocol involves extra influence factors such as simulation parameters. To exclude irrelevant factors, we choose to randomly augment the appearance of the conveyor belt in Isaac Gym, as visualized in Fig.~\ref{Fig: diverse_backgrounds}. In this way, the policy is trained in Isaac Gym and validated in Isaac Gym, where the environment background is the only difference. We evaluate the performance of GEM and present the results in Table~\ref{Table: environment generalization study}.

According to the results, the performances of GEM are similar in environments with seen and unseen backgrounds, which is consistent with the expectation. In addition, the subsequent sim-to-real experiment in Table~\ref{Table: sim-to-real generalization} indicates that the GEM trained using only simulated data can generalize to real-world manipulation despite the appearance gap between the simulated and real-world environments.

\begin{figure}[tbp]
    \centering
    \includegraphics[width=1.0\linewidth]{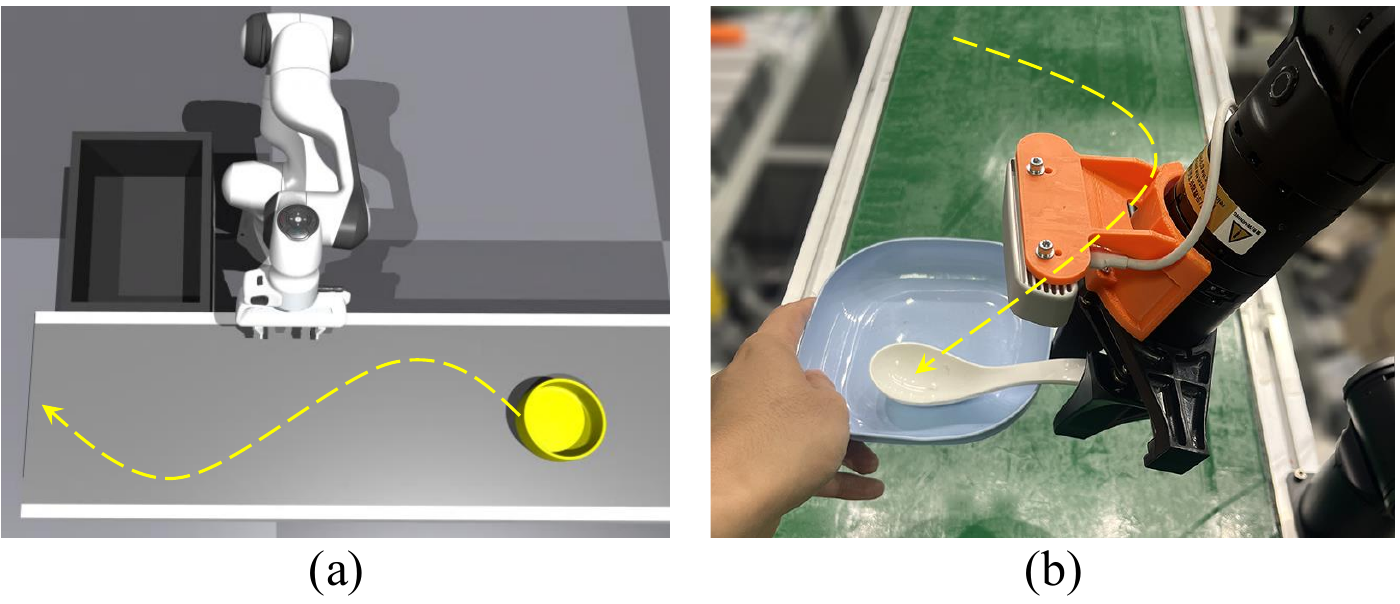}
    \caption{This figure illustrates how we evaluate the motion pattern generalization capability of GEM in simulated and real environments. Different from the linear motion in demonstrations, in subfigure (a), the object moves following an S-shaped trajectory. In subfigure (b), the object is manually held and moves along a random curved trajectory.} \label{Fig: motion_patterns}
\end{figure}

\begin{table}[tbp] 
    \centering
    \caption{Generalization study on motion patterns.} \label{Table: motion pattern study}
    \vspace{-0.2cm}
    \tabcolsep=0.17cm
    \resizebox{1.0\linewidth}{!}{
    \begin{tabular}{cc|cccc}
    \toprule
    Environment & Motion & Pick$\uparrow$ & Put$\uparrow$ & Rotate$\uparrow$ & Insert$\uparrow$ \\
    \midrule
    Simulated & Linear & 0.94 & 1.00 & 0.90 & 0.86 \\
    Simulated & S-curve & 0.95 & 1.00 & 0.88 & 0.81 \\
    Real & Linear & 0.98 & 1.00 & 0.93 & 0.89 \\
    Real & Random curve & 0.97 & 1.00 & 0.90 & 0.75 \\
    \bottomrule
    \end{tabular}}
\end{table}

\subsubsection{Generalization across Motion Patterns}
As the tracking actions generated based on the manipulation action decomposition strategy transform the original dynamic object manipulation problem into a static object manipulation problem, the GEM policy should be able to generalize to manipulating objects in diverse motion patterns, \eg, various moving velocities and trajectories. The experimental results in Table~\ref{Table: speed analysis} have confirmed that the manipulation success rates remain similar when the conveyor belt speed is adjusted but does not surpass the robot hand tracking velocity limit. In this experiment, we further study this problem by altering the object moving trajectories. Specifically, in simulation, we program the objects to move on the conveyor belt following an S-shaped trajectory as illustrated in Fig.~\ref{Fig: motion_patterns}~(a). In real-world evaluation, we manually hold a container and put the object to manipulate in this container, moving it along a random curved trajectory, as shown in Fig.~\ref{Fig: motion_patterns}~(b).

We report the experiment results in Table~\ref{Table: motion pattern study}. Comparing the performance between linear and curved object movement trajectories, we can find that making objects move along an unknown curved trajectory does not harm the manipulation success rates significantly, especially for the tasks that do not require precise operation like Put. Therefore, GEM is robust to different motion patterns of moving objects. 

\begin{figure}[t]
    \centering
    \includegraphics[width=1.0\linewidth]{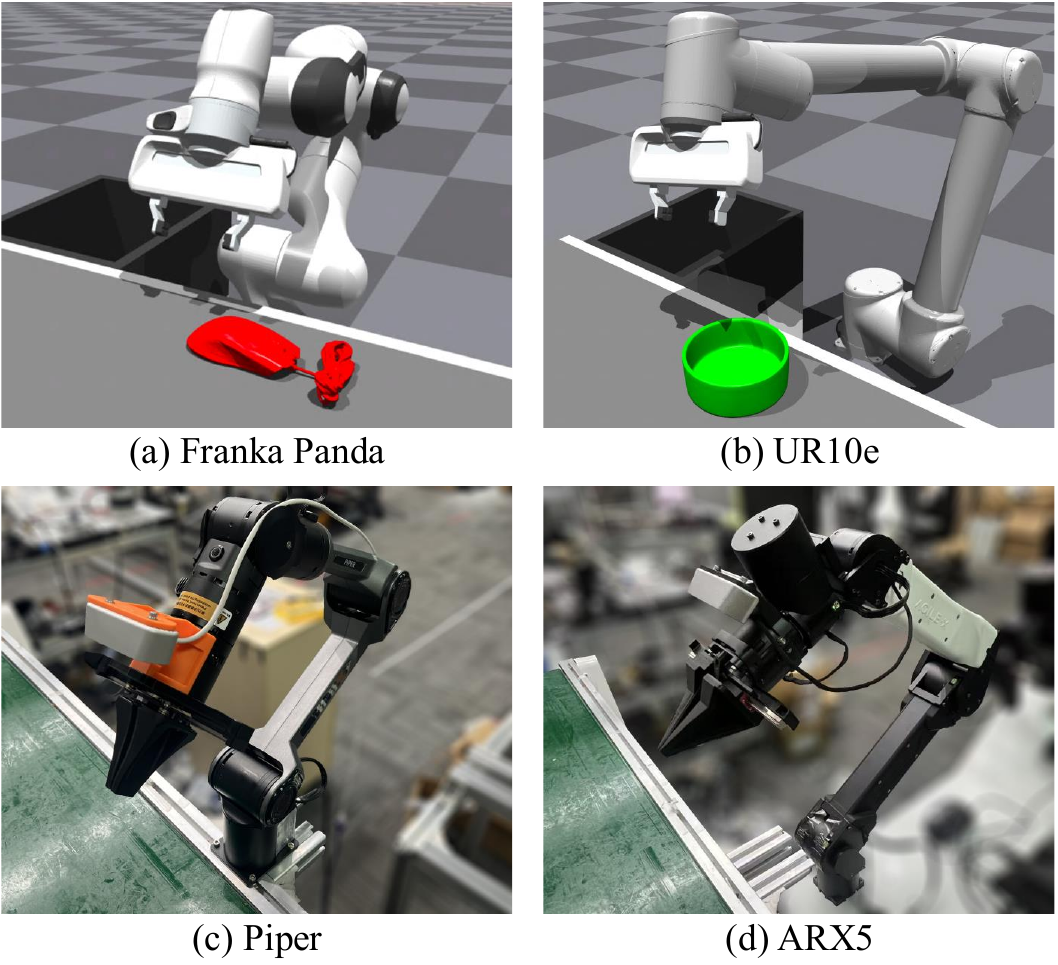}
    \caption{This figure visualizes the four kinds of robot hands to conduct cross-embodiment evaluation in the simulated and real environments.} \label{Fig: robot_embodiments}
\end{figure}

\begin{table}[t] 
    \centering
    \caption{Generalization study on unseen objects.} \label{Table: object geometry study}
    \vspace{-0.2cm}
    \tabcolsep=0.2cm
    \resizebox{0.85\linewidth}{!}{
    \begin{tabular}{c|ccccc}
    \toprule
    Object & Pick$\uparrow$ & Put$\uparrow$ & Rotate$\uparrow$ & Insert$\uparrow$ & Avg$\uparrow$ \\
    \midrule
    Seen & 0.94 & 1.00 & 0.90 & 0.86 & 0.93 \\
    Unseen & 0.89 & 0.98 & 0.87 & 0.85 & 0.90 \\
    \bottomrule
    \end{tabular}}
\end{table}

\subsubsection{Generalization across Unseen Objects}
Besides environment backgrounds and object motion patterns, generalizing to manipulating unseen objects is also important in real application scenarios. The experimental results reported in Table~\ref{Table: Simulation Comparison} and Table~\ref{Table: Real Comparison} have suggested that the GEM policy can operate novel objects. This advantage is thanks to the training demonstrations collected on objects of diverse geometries, enabling the GEM network to generalize to out-of-domain object geometries. However, there exist extra influence factors in these two experiments, such as inconsistent simulation parameters and new robot embodiments. In this experiment, we further analyze the performance of GEM on manipulating unseen objects after excluding irrelevant factors.  

To this end, we import the novel 3D assets shown in Fig.~\ref{Fig: IsaacSim_envi} (b) into Isaac Gym to evaluate the GEM trained based on the demonstrations from Isaac Gym. The experiment results are reported in Table~\ref{Table: object geometry study}. We can observe that the success rates of GEM in manipulating out-of-domain objects are similar to the ones of operating in-domain objects. This is because GEM primarily exploits geometry information, and the basic 3D geometries of objects are limited despite their diverse visual appearances. The 3D assets depicted in Fig.~\ref{Fig: simulated_envi}~(a) have covered many common basic geometries.

\subsubsection{Generalization across Robot Embodiments}
Because of the developed manipulation action decomposition strategy, the end-effector moves to a preset position relative to the target object before the predicted interaction actions are applied. Additionally, the manipulation actions control the joints of a robot based on inverse kinematics. Therefore, GEM is not restricted to a single robot embodiment. To evaluate the robot embodiment generalization, we conduct experiments using four kinds of robot hands in simulated and real-world environments, as shown in Fig.~\ref{Fig: robot_embodiments}. In the simulation based on Isaac Gym, the policy trained using the simulated demonstrations collected from Franka Panda is directly applied to the UR10e robot hand. In the real-robot experiment, the policy tuned based on the demonstrations collected with Piper is deployed on the ARX5 robot hand to conduct manipulation. 

The experiment results are reported in Table~\ref{Table: robot embodiment study}. It is observed that employing a robot to collect demonstrations and adopting another robot to conduct manipulation evaluation achieves similar performance compared with the setting where the demonstration collection and validation are based on the same robot. Hence, the robot embodiment generalization capability of GEM is confirmed.

\begin{figure}[t]
    \centering
    \includegraphics[width=1.0\linewidth]{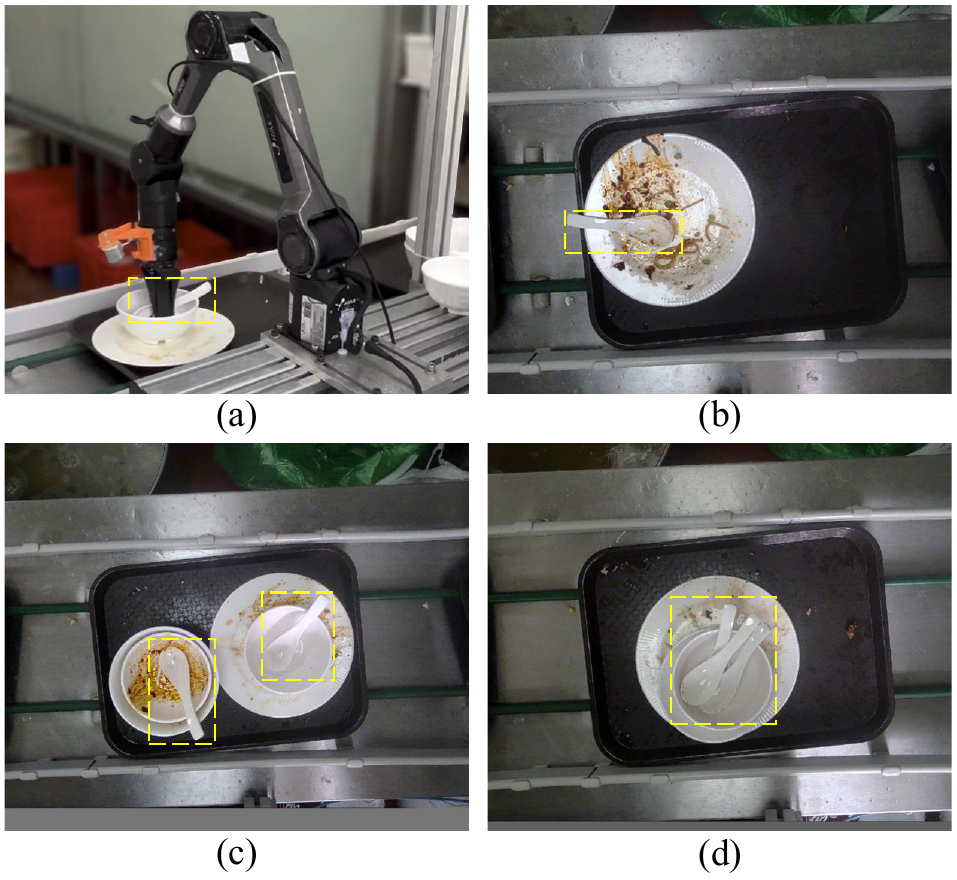}
    \caption{Subfigure (a) shows the real application, where the robot hand needs to collect spoons from the trays moving on a conveyor belt. Subfigures (b)-(d) illustrate some challenging cases, \ie, spoons with food residue, multiple spoons to pick up within a short time window, and spoons stacked together.} \label{Fig: application_scene}
\end{figure}

\begin{table}[t] 
    \centering
    \caption{Generalization study on robot embodiments.} \label{Table: robot embodiment study}
    \vspace{-0.2cm}
    \tabcolsep=0.2cm
    \resizebox{0.9\linewidth}{!}{
    \begin{tabular}{cc|cccc}
    \toprule
    Train & Validation & Pick$\uparrow$ & Put$\uparrow$ & Rotate$\uparrow$ & Insert$\uparrow$ \\
    \midrule
    Franka & Franka & 0.94 & 1.00 & 0.90 & 0.86 \\
    Franka & UR10e & 0.95 & 1.00 & 0.88 & 0.86 \\
    Piper & Piper & 0.98 & 1.00 & 0.93 & 0.89 \\
    Piper & ARX5 & 0.96 & 1.00 & 0.94 & 0.85 \\
    \bottomrule
    \end{tabular}}
    \vspace{-0.2cm}
\end{table}

\begin{table}[t] 
    \centering
    \caption{Sim-to-real performance of GEM.} \label{Table: sim-to-real generalization}
    \vspace{-0.2cm}
    \tabcolsep=0.25cm
    \resizebox{0.95\linewidth}{!}{
    \begin{tabular}{c|ccccc}
    \toprule
    Environment & Pick$\uparrow$ & Put$\uparrow$ & Rotate$\uparrow$ & Insert$\uparrow$ & Avg$\uparrow$ \\
    \midrule
    Real\_tune & 0.98 & 1.00 & 0.93 & 0.89 & 0.95 \\
    Real\_zero & 0.91 & 0.98 & 0.80 & 0.79 & 0.87 \\
    \bottomrule
    \end{tabular}}
\end{table}

\subsubsection{Sim-to-Real Generalization}
According to the aforementioned experiments, GEM can generalize to novel environment backgrounds, motion patterns, target objects, and robot embodiments. These characteristics enable the possibility of generalizing to real-world tasks based on simulated demonstrations. To validate this issue, we design two real-robot experiment settings, \ie, tuned sim-to-real and zero-shot sim-to-real. The experimental results of the tuned sim-to-real setting are presented in Table~\ref{Table: Real Comparison}. The results suggest that GEM achieves significantly superior sim-to-real generalization with a handful of tuning demonstrations.

The performance of GEM in the zero-shot sim-to-real setting is reported in Table~\ref{Table: sim-to-real generalization}. According to the results, it is observed that the GEM policy trained using only simulated data can manipulate objects in the real world, and its success rates are still promising. This strong generalization is attributed to the design that GEM prioritizes geometry information and decreases the importance of visual appearance. We also analyze the reasons of the slight performance deterioration. An important factor is there exists a control execution delay between the input signal and the robot system response. This control execution delays are different between simulated and real robots, causing a sim-to-real execution gap. Tuning the policy with a few real-robot demonstrations can bridge this execution delay gap. Therefore, the performance of GEM in the tuned sim-to-real setting is stronger.

\subsection{Real Application Deployment}
\label{sub:real_application_deployment}

Different from lab environments, it is often infeasible to collect demonstrations in real application scenes because of the uninterrupted normal workflows and strict confidentiality requirements. Therefore, if a policy is to be deployed, it should be capable of operating in unseen environments with consistently high success rates, handling unexpected object poses and lighting changes. The aforementioned techniques proposed in this work help GEM meet these demands.

To evaluate the robustness of GEM, we deploy it in a canteen kitchen to conduct tableware collection. Specifically, after dining, canteen customers place their used tableware on a conveyor belt, which transports these items to the kitchen. The manipulation system needs to identify and pick out the spoons from the tableware and place them into a separate container, as shown in Fig.~\ref{Fig: application_scene}~(a). The spoons are placed in varying poses and may shift during transport. In addition, as depicted in Fig.~\ref{Fig: application_scene} (b)-(d), other challenges include food residue obscuring spoon pose recognition, the robot arm needing to grasp and place multiple spoons within tight time windows, and spoons stacked together, \etc.

Since onsite data collection is infeasible in this canteen, we teleoperate 100 demonstrations using the manipulation platform shown in Fig.~\ref{Fig: real_robot}. Then, we utilize these 100 demonstrations to fine-tune the GEM policy that has been pre-trained with the simulation demonstrations. Finally, we deploy the fine-tuned GEM in this canteen to conduct spoon collection. We evaluate GEM over a seven-day period, processing more than ten thousand spoons, and measure a final manipulation success rate of 97.2\%.

\section{Conclusion}
\label{sec:conclusion}

This paper has first pointed out that embodied action generation concerns more about the 3D geometry structure of environments than visual appearance. Based on this insight, we have proposed the policy GEM for dynamic object manipulation on conveyor systems. The key design principle of this policy is prioritizing the exploitation of 3D geometry information and decreasing the reliance on 2D visual information. In addition, we have developed simulation platforms of the explored tasks to generate training demonstrations efficiently. Since the simulators can simulate the geometry-based interaction dynamics among objects well and GEM primarily uses geometry information, GEM achieves promising sim-to-real generalization. We have conducted extensive experiments to analyze the characteristics of GEM. The results indicate that GEM has successfully generalized across diverse environment backgrounds, motion patterns, object geometries, and robot embodiments, achieving a success rate of over 97\% in a real-world application without using any in-scene demonstration. These results have demonstrated that our policy design principle effectively guides the design of an efficient manipulation system, offering a practical solution for industrial automation.

{\small
\bibliographystyle{ieeetr}
\bibliography{reference}
}

\end{document}